\documentclass[pmlr,twoside,10pt]{jmlr}

\jmlrvolume{}
\jmlryear{}
\jmlrproceedings{}{}
\jmlrpages{}

\makeatletter
\def\@titlefoot{}
\makeatother

\usepackage{tikz}
\usetikzlibrary{shapes, arrows.meta, positioning}
\makeatletter
\@for\next:={For,EndFor,ForAll,ForEach,If,ElsIf,Else,EndIf,While,EndWhile,%
Repeat,Until,Loop,EndLoop,Function,EndFunction,Procedure,EndProcedure,%
Return,Require,Ensure,State,Statex,Comment,Call}\do{%
  \expandafter\let\csname\next\endcsname\relax}
\makeatother
\usepackage{algorithm}
\usepackage{algpseudocode}
\usepackage{float}
\usepackage{comment}

\definecolor{iitbblue}{RGB}{0, 62, 116}
\definecolor{resultgreen}{RGB}{34, 139, 34}
\definecolor{highlight}{RGB}{204, 85, 0}

\theorembodyfont{\upshape}

\DeclareMathOperator*{\argmax}{arg\,max}

\title[Local Indices Are Not Enough: Synchrony and Global
Information in Bandit Algorithms]{Local Indices Are Not Enough: Synchrony and Global
Information in Bandit Algorithms}

\author{%
 \Name{Aakash Gore} \Email{aakash.gore@iitb.ac.in}\\
 \addr Department of Electrical Engineering,
       Indian Institute of Technology Bombay
 \AND
 \Name{Shivaram Kalyanakrishnan} \Email{shivaram@cse.iitb.ac.in}\\
 \addr Department of Computer Science,
       Indian Institute of Technology Bombay
}

\begin{document}

\maketitle

\begin{abstract}
Standard index-based stochastic bandit algorithms such as KL-UCB and Thompson Sampling achieve
optimal $O(\log T)$ regret over a horizon $T$. The former uses the {\it global} timestep to construct each arm's index, while the latter samples {\it each} arm's belief distribution at {\it every} timestep. The necessity of global information and synchronous index updates for achieving optimal regret has not been examined yet. In this paper, we investigate the fundamental
limits of \emph{asynchronous local} index algorithms---those that recalculate an arm's
index only upon pulling it, using solely that arm's own reward history and a local
randomness budget. We prove that no such
algorithm can achieve $O(\log T)$ regret across all bandit instances. We
then analyse the functional equations arising from asynchronous updates, and show that the achievable regret lies in
a sub-polynomial, super-polylogarithmic regime. We propose an asynchronous local index
policy whose regret upper bound also lies in this regime. Turning to \emph{synchronous local} algorithms, which sample {\it every} arm's
index at {\it all} timesteps as in Thompson Sampling, we show that a {\it locally determined} randomness
budget per arm suffices to retain $O(\log T)$ regret, demonstrating that synchrony, albeit with {\it some} randomness, is essential to the optimality of Thompson Sampling.
\end{abstract}


\section{Introduction}


Multi-armed bandits \citep{thompson1933,robbins1952} are a fundamental problem in the sequential decision-making framework, where an agent sequentially chooses an action (pulls an arm) and receives a reward from the environment. In the stochastic bandit setting, each action is associated with an unknown probability distribution from which rewards are independently sampled. The central challenge in this framework is the exploration--exploitation tradeoff: the agent must explore to identify high-reward actions while increasingly exploiting the optimal action to maximize cumulative payoff. The algorithm governing this behavior is referred to as the learning policy, and its performance is quantified by its regret, defined as the expected difference between the cumulative reward of the policy which always pulls the best arm and that of the learning policy over a horizon $T$.

\paragraph{The Bandit Instance.}
Let $\mathcal{F}$ denote the set of all univariate probability distributions supported on $[0,1]$. A \emph{stochastic bandit instance} is a triple
\[
\nu \;=\; \big(B,\ (\nu_b)_{b \in B},\ \ell\big),
\]
where $B$ is a finite set of \emph{arms} with $K := |B| \ge 2$, each $\nu_b \in \mathcal{F}$ is the reward distribution of arm $b$, and $\ell : B \to \mathbb{N}$ is an injective \emph{labelling}; we write $L := \ell(B)$ for the instance's label set. We distinguish an arm, the physical object carrying a distribution, from its label, the name the algorithm uses to address it. 

For each arm $a$, let $\mu_a = \mathbb{E}_{X \sim \nu_a}[X]$ denote its expected reward. Let $a^\ast$ denote an optimal arm (ties broken arbitrarily), with corresponding mean $\mu^\ast = \max_a \mu_a$, and let $\Delta_a = \mu^\ast - \mu_a$ denote the suboptimality gap of arm $a$.

\paragraph{The Algorithm.}
Interaction proceeds in discrete global timesteps $t = 1, 2, \dots, T$. At each timestep $t$, the algorithm selects an arm $A_t$ and observes a reward $X_{A_t, t} \sim \nu_{A_t}$, drawn independently of the past. Let $\mathcal{H}^{t-1} = \sigma(A_1, X_{A_1, 1}, \ldots, A_{t-1}, X_{A_{t-1}, t-1})$ denote the $\sigma$-algebra representing the global history up to time $t-1$, and let $\mathcal{S}_K$ denote the probability simplex over the arms. A general algorithm $\pi$ is a sequence of mappings $(\pi_1, \pi_2, \dots)$ where each $\pi_t : \mathcal{H}^{t-1} \to \mathcal{S}_K$ maps the global history to a distribution over the $K$ arms.

The regret incurred by an algorithm $\pi$ is defined as
\[
R_{T}(\pi,\nu) = \mathbb{E}_{\pi, \nu}\left[ \sum_{t=1}^T \left( \mu^\ast - X_{A_t, t} \right) \right] = \sum_{a \neq a^\ast} \Delta_a \, \mathbb{E}_{\pi, \nu}[N_a(T)],
\]
where $N_a(T) = \sum_{t=1}^T \mathbb{I}\{A_t = a\}$ denotes the number of times arm $a$ is pulled up to time $T$.

It is well known that any uniformly good algorithm---one that achieves sub-polynomial regret on all bandit instances---must incur at least logarithmic regret asymptotically on all instances. By Theorems 1 and 2 of \citet{lai1985asymptotically}, if an algorithm $\pi$ incurs regret $o(T^\alpha)$ for all $\alpha > 0$ on all instances, then on any instance $\nu$ whose arms belong to a one-parameter family $\{f(\cdot;\theta)\}_{\theta\in\Theta}\subset\mathcal{F}$ satisfying conditions (1.6), (1.7) and (1.9) of \citet{lai1985asymptotically}, and for every suboptimal arm $a$,
\[
\liminf_{T \rightarrow \infty} \frac{R_{T}(\pi,\nu)}{\log T} \geq C_{\nu},
\qquad\text{and}\qquad
\lim_{T\rightarrow\infty} \mathbb{P}\left(N_a(T) \geq \frac{(1-\epsilon)\log T}{D_{\nu,a}}\right) = 1 \quad \forall \epsilon > 0,
\]
where $C_{\nu}$ is an instance-dependent constant and $D_{\nu,a} := I(\theta_a,\theta^\ast) \in (0,\infty)$ is the Kullback--Leibler divergence between the reward distributions of $a$ and of the optimal arm.

Many bandit algorithms are \emph{index-based}: at every timestep each arm $a$ is summarised by a
scalar index $I_a(t)$, computed from the history so far, and the algorithm pulls an arm with the
largest index. An index typically combines the arm's empirical mean with a term that favours arms
pulled less often, so that every arm keeps being revisited.
A large body of work has developed algorithms whose upper bounds match this $\Omega(\log T)$ lower bound, but each of them uses information beyond a single arm's own reward history. UCB \citep{auer2002finite}, KL-UCB \citep{garivier2011kl} and IMED \citep{honda2015imed} are deterministic, and their indices depend explicitly on global quantities: the current timestep $t$ for the first two, the empirically best arm for the third. Thompson Sampling \citep{thompson1933} is randomized and does construct each arm's belief from that arm's local reward history alone; however, it resamples every arm's index at every timestep, regardless of which arm was selected.

These two departures from locality are different in kind, and we name them separately. An algorithm is \emph{synchronous} if it updates all arm indices at every timestep, and \emph{asynchronous} if it updates an arm's index only when that arm is pulled. Independently of this, an arm's index is \emph{global} if it depends on the horizon, the current timestep, or statistics of other arms' histories, and \emph{local} if it depends only on that arm's own history. Thus KL-UCB is global, while Thompson Sampling is local but synchronous. Section~\ref{sec:classes} makes both notions precise.

In this work we study the limits of asynchronous local index algorithms, which have neither
ingredient: they read no global quantity, and they recompute an arm's index only when that arm is pulled.
We prove that no algorithm in this class achieves polylogarithmic regret on all instances, describe a set of regret rates the class can attain, and construct a policy attaining them. Since KL-UCB reads the global
clock and Thompson Sampling resamples every arm at every round, each optimal algorithm uses an ingredient
this class lacks, and our results show that one of the two is indispensable.

\subsection{Algorithm Classes}
\label{sec:classes}

We define the local information set available to each arm, and formalize the algorithm classes we study in this work.

\paragraph{Local Information.}
For any arm $a$, we define its \emph{local information} at time $t$ as $L_a^t = (N_a(t), \mathcal{H}_a^t)$, where $\mathcal{H}_a^t = (X_{a,1}, \dots, X_{a,N_a(t)})$ is the sequence of rewards obtained from pulls of arm $a$. Thus the local information of an arm captures everything it can infer about itself, without access to the global clock or the histories of other arms; in particular, $L_a^t$ contains neither the global time $t$ nor the horizon $T$.

\medskip
\noindent\textbf{Example.} For a two-armed bandit with interaction sequence
\[
\{(A_t, X_{A_t,t})\}_{t=1}^{5} = \{(1,0.1),\ (2,0.6),\ (2,0.3),\ (1,0.2),\ (2,0.8)\},
\]
the local histories are
\[
\mathcal{H}_1^{5} = (0.1,\,0.2),\quad N_1(5)=2,
\qquad
\mathcal{H}_2^{5} = (0.6,\,0.3,\,0.8),\quad N_2(5)=3.
\]
These retain only the rewards obtained from each arm, discarding the timesteps at which they were observed and any information about the other arm.

\begin{figure}[t]
\centering

\begin{minipage}[t]{0.48\textwidth}
\centering
\begin{tikzpicture}[
    baseline=(agg.north),
    arm/.style={draw, rounded corners=5pt, minimum width=1.9cm, minimum height=0.9cm,
                align=center, font=\footnotesize},
    >=Stealth
]
    \path (-3.6,-0.95) rectangle (3.6,3.05);
    \node[draw, rounded corners=5pt, minimum width=3.0cm, minimum height=1.1cm,
          fill=iitbblue!10, font=\footnotesize, align=center]
        (agg) at (0, 2.4) {\textbf{Aggregator}\\[-2pt]
        {\scriptsize $g_{\mathrm{tie}}(\arg\max_a I_a)$}};

    \node[arm, fill=resultgreen!15] (a1) at (-2.2, 0) {\textbf{Arm 1}\\[-2pt]{\scriptsize $I_1{=}f_{\mathrm{async}}(1,L_1^t,S_1)$}};
    \node[arm, fill=gray!8] (a2) at (2.2, 0) {\textbf{Arm 2}\\[-2pt]{\scriptsize $I_2{=}f_{\mathrm{async}}(2,L_2^t,S_2)$}};

    \draw[->, gray!70] (a1.north) to[bend left=15]
        node[pos=0.28, left, font=\tiny, text=gray]{$I_1$} (agg.south west);
    \draw[thick, resultgreen, ->] (agg.south west) to[bend left=15]
        node[pos=0.72, right, font=\tiny, text=resultgreen]{pull} (a1.north);

    \draw[->, gray!70] (a2.north) --
        node[midway, right, font=\tiny, text=gray]{$I_2$} (agg.south east);

    \draw[thick, dashed, gray!50] (0, -0.8) -- (0, 0.8);

\end{tikzpicture}

\vspace{2pt}
{\small\raggedright\textbf{(a)} Local ($\mathcal{L}_{\mathrm{Async}}$, $\mathcal{L}_{\mathrm{Sync}}$). Each arm
computes $I_a$ from its own local information $L_a^t=(N_a(t),\mathcal{H}_a^t)$ only, with no access
to the global time or to the other arm's history; the dashed line marks this isolation.\par}
\label{fig:local-arch}
\end{minipage}
\hfill
\begin{minipage}[t]{0.48\textwidth}
\centering
\begin{tikzpicture}[
    baseline=(agg.north),
    arm/.style={draw, rounded corners=5pt, minimum width=1.9cm, minimum height=0.9cm,
                align=center, font=\footnotesize},
    >=Stealth
]
    \path (-3.6,-0.95) rectangle (3.6,3.05);
    \node[draw, rounded corners=5pt, minimum width=3.0cm, minimum height=1.1cm,
          fill=iitbblue!15, thick, font=\footnotesize, align=center]
        (agg) at (0, 2.4) {\textbf{Aggregator}\\[-2pt]
        {\scriptsize computes all $I_a(t){=}f(a,\mathcal{H}^t)$}};

    \node[arm, fill=blue!8] (a1) at (-2.2, 0) {\textbf{Arm 1}};
    \node[arm, fill=blue!8] (a2) at (2.2, 0) {\textbf{Arm 2}};

    \draw[->, gray!70] (a1.north) to[bend left=15]
        node[pos=0.28, left, font=\tiny, text=gray]{$\mathcal{H}_1^t$} (agg.south west);
    \draw[thick, resultgreen, ->] (agg.south west) to[bend left=15]
        node[pos=0.72, right, font=\tiny, text=resultgreen]{pull} (a1.north);

    \draw[->, gray!70] (a2.north) --
        node[midway, right, font=\tiny, text=gray]{$\mathcal{H}_2^t$} (agg.south east);

    \node[font=\tiny, text=gray, rotate=90] at (0, 0) {\it no wall};

\end{tikzpicture}

\vspace{2pt}
{\small\raggedright\textbf{(b)} Global (e.g.\ KL-UCB, IMED). Every index is computed from the joint history
$\mathcal{H}^t=(\mathcal{H}_1^t,\mathcal{H}_2^t)$, which records when each arm was pulled and so
reveals the global time $t$ even without an explicit clock.\par}
\label{fig:global-arch}
\end{minipage}

\caption{Local versus global index architectures. In (a) the aggregator receives only the
indices, never the arm histories or a global clock, and pulls an arm with the largest index,
breaking ties with $g_{\mathrm{tie}}$. $\mathcal{L}_{\mathrm{Async}}$ and
$\mathcal{L}_{\mathrm{Sync}}$ (Table~\ref{tab:algos}) are both of type (a); they differ only in
\emph{when} each arm's index is recomputed.}
\label{fig:local-vs-global}
\end{figure}
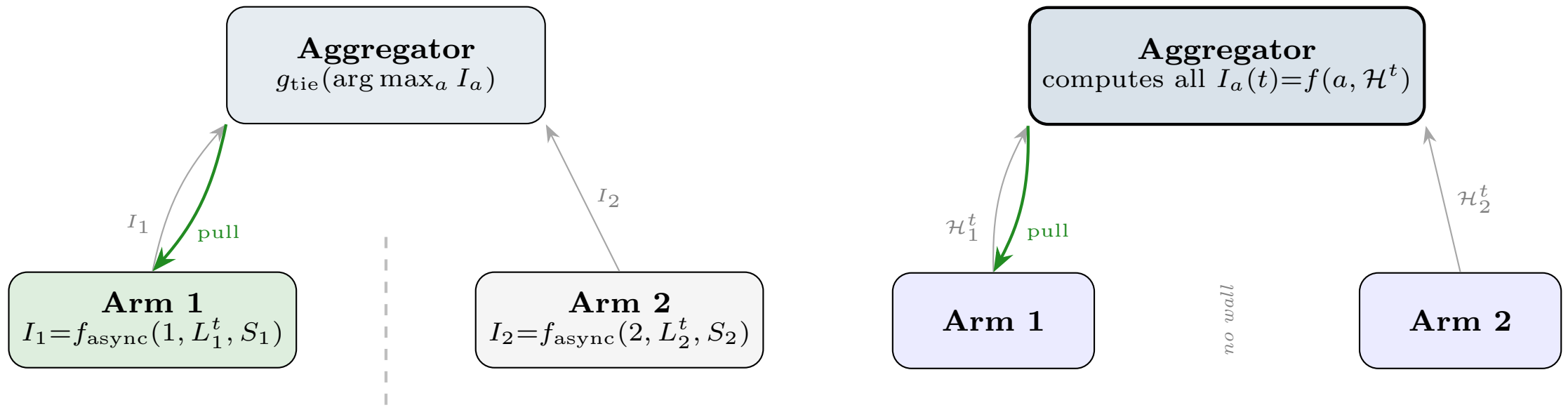

\medskip
It is useful to picture a central \emph{aggregator} which expects each arm $a$ to furnish an index $I_a(t)$ at timestep $t$, and then pulls the arm with the maximum index. If the algorithm is global, each arm can \emph{see} the other arms, i.e.\ access their histories; if it is local, the arms cannot see each other and may access only their own histories. In either case the histories are abstracted away from the aggregator, which observes only the indices furnished to it. Figure~\ref{fig:local-vs-global} contrasts the two architectures.

\paragraph{Tie-Breaking.}
Several arms may share the same maximal index, so the aggregator needs a rule for choosing among them. Since the aggregator observes only the furnished indices, such a rule can depend on nothing but the set of tied labels.

\begin{definition}[Deterministic Tie-Breaking]
\label{def:tiebreaking}
A \emph{deterministic tie-breaking rule} is a map $g_{\mathrm{tie}}$ assigning to each non-empty finite set $\Omega \subset \mathbb{N}$ of labels an element $g_{\mathrm{tie}}(\Omega) \in \Omega$.
\end{definition}

Given the set of index-maximizing labels $\Omega(t) = \argmax_{a \in L} I_a(t)$, the aggregator selects
\[
A_{t+1} = g_{\mathrm{tie}}\big(\Omega(t)\big).
\]
Definition~\ref{def:tiebreaking} asks only that $g_{\mathrm{tie}}$ be a choice function: its value on $\Omega$ is determined by $\Omega$ alone, but no relation whatsoever is imposed between its values on different sets. Randomized tie-breaking is addressed in Remark~\ref{rem:randomized_tie} of Section~\ref{sec:embedding}.

\begin{definition}[Asynchronous Local Index Algorithms, $\mathcal{L}_{\mathrm{Async}}$]
\label{def:async}
An algorithm $\pi$ belongs to class $\mathcal{L}_{\mathrm{Async}}$ if there is a function $f_{\mathrm{async}}$ such that for each arm $a$, the index at time $t$ is given by
\[
I_a(t) = f_{\mathrm{async}}\big(a, L_a^t, S_a^{N_a(t)}\big),
\]
where $S_a^{N_a(t)}$ is a random bit string of length $\rho(L_a^t)$ (possibly infinite), and $\rho$ is a mapping from local histories to $\{0,1,2,\dots\}\cup\{\infty\}$.

\end{definition}
Note that, unless arm $a$ was pulled in the previous round, $L_a^t$ and $S_a^{N_a(t)}$ don't change, and hence the index $I_a(t)$ also does not change. Hence, the index of arm $a$ is updated only when arm $a$ is pulled, and remains unchanged otherwise.

\begin{definition}[Synchronous Local Index Algorithms, $\mathcal{L}_{\mathrm{Sync}}$]
\label{def:sync}
An algorithm $\pi$ belongs to class $\mathcal{L}_{\mathrm{Sync}}$ if there is a function $f_{\mathrm{sync}}$ such that for each arm $a$, the index at time $t$ is given by
\[
I_a(t) = f_{\mathrm{sync}}\big(a, L_a^t, S_a^{t}\big),
\]
where $S_a^{t}$ is a random bit string of length $\rho(L_a^t) \geq 1$ (possibly infinite), and $\rho$ is a mapping from local histories to $\{1,2,\dots\}\cup\{\infty\}$.  

\end{definition}
Note that at each timestep, $S_a^t$ is resampled regardless of whether arm $a$ was pulled in the previous timestep or not. Our results concern $\mathcal{L}_{\mathrm{Async}}$; we define $\mathcal{L}_{\mathrm{Sync}}$ only as a point of comparison, since Thompson Sampling belongs to it and is optimal \citep{agrawal2012analysis,kaufmann2012thompson}.

\textbf{The key distinction is sampling}. Asynchronous Local Index algorithms update an arm's index only upon pulling it, whereas Synchronous Local Index algorithms recompute each arm's index at every global timestep. Table~\ref{tab:algos} presents both classes side by side from the aggregator's point of view. In both cases, the aggregator maintains an index array, selects an arm via $g_{\mathrm{tie}}$, and observes a reward. The classes differ in what happens next: $\mathcal{L}_{\mathrm{Async}}$ recomputes only the pulled arm's index (via $f_{\mathrm{async}}$), while $\mathcal{L}_{\mathrm{Sync}}$ recomputes every arm's index (via $f_{\mathrm{sync}}$).

\begin{table}[h]
\centering
\caption{Aggregator loop for asynchronous vs.\ synchronous local index algorithms. Both columns show the same aggregator stepping through time: it holds the index array, selects via $g_{\mathrm{tie}}$, and observes a reward. They differ only in the \textbf{index update} phase (highlighted): $\mathcal{L}_{\mathrm{Async}}$ resamples and recomputes the pulled arm's index via $f_{\mathrm{async}}$ while freezing all others; $\mathcal{L}_{\mathrm{Sync}}$ resamples and recomputes \emph{every} arm's index via $f_{\mathrm{sync}}$.}
\label{tab:algos}
\renewcommand{\arraystretch}{1.1}
\begin{tabular}{@{}p{0.48\textwidth}@{\hspace{0.5em}}|@{\hspace{0.5em}}p{0.48\textwidth}@{}}
\multicolumn{1}{c}{\textbf{$\mathcal{L}_{\mathrm{Async}}$ (Asynchronous)}} &
\multicolumn{1}{c}{\textbf{$\mathcal{L}_{\mathrm{Sync}}$ (Synchronous)}} \\
\hline
\\[-0.7em]
\begin{minipage}[t]{0.48\textwidth}
\vspace{0pt}
\begin{algorithmic}[1]
\State \textbf{Input:} Arms $[K]$, index $f_{\mathrm{async}}$, rule $g_{\mathrm{tie}}$
\State \textbf{Init:} $N_a(0) \gets 0$; sample $S_a^0$
\State \quad $I_a(0) \gets f_{\mathrm{async}}(a, L_a^0, S_a^0)$ $\forall a$
\For{$t = 0, 1, \dots, T-1$}
    \State $A_{t+1} \gets g_{\mathrm{tie}}(\arg\max_{a} I_a(t))$
    \State Observe $X_{A_{t+1},t+1}$
    \State $L_{A_{t+1}}^{t+1} \gets \mathrm{update}(L_{A_{t+1}}^t, X_{A_{t+1},t+1})$
    \State $L_a^{t+1} \gets L_a^t$ for $a \neq A_{t+1}$
    \State {\color{blue}\textbf{Update pulled arm only:}}
    \State \quad {\color{blue}Sample $S_{A_{t+1}}^{N_{A_{t+1}}(t+1)}$}
    \State \quad {\color{blue}$I_{A_{t+1}}(t{+}1) \gets f_{\mathrm{async}}(A_{t+1}, L_{A_{t+1}}^{t+1}, S_{A_{t+1}}^{N_{A_{t+1}}})$}
    \State {\color{blue}\textbf{Freeze all other arms:}}
    \State \quad {\color{blue}$I_a(t{+}1) \gets I_a(t)$ for $a \neq A_{t+1}$}
\EndFor
\end{algorithmic}
\end{minipage}
&
\begin{minipage}[t]{0.48\textwidth}
\vspace{0pt}
\begin{algorithmic}[1]
\State \textbf{Input:} Arms $[K]$, index $f_{\mathrm{sync}}$, rule $g_{\mathrm{tie}}$
\State \textbf{Init:} $N_a(0) \gets 0$; sample $S_a^0$
\State \quad $I_a(0) \gets f_{\mathrm{sync}}(a, L_a^0, S_a^0)$ $\forall a$
\For{$t = 0, 1, \dots, T-1$}
    \State $A_{t+1} \gets g_{\mathrm{tie}}(\arg\max_{a} I_a(t))$
    \State Observe $X_{A_{t+1},t+1}$
    \State $L_{A_{t+1}}^{t+1} \gets \mathrm{update}(L_{A_{t+1}}^t, X_{A_{t+1},t+1})$
    \State $L_a^{t+1} \gets L_a^t$ for $a \neq A_{t+1}$
    \State {\color{red}\textbf{Resample all indices:}}
    \For{$a \in [K]$}
        \State \quad {\color{red}Sample $S_a^{t+1}$}
        \State \quad {\color{red}$I_a(t{+}1) \gets f_{\mathrm{sync}}(a, L_a^{t+1}, S_a^{t+1})$}
    \EndFor
\EndFor
\end{algorithmic}
\end{minipage}
\\
\end{tabular}
\end{table}
\subsection{Our Results}


\paragraph {$\boldsymbol{\mathcal{L}_{\mathrm{Async}}}$ Algorithms cannot achieve polylogarithmic regret.}
We ask whether any asynchronous local index algorithm can achieve $O(\log T)$ regret. Using a novel \emph{pathwise embedding argument}, we prove in Section~\ref{sec:embedding} that the answer is no: for every $\mathcal{A} \in \mathcal{L}_{\mathrm{Async}}$ there is an instance on which the regret of $\mathcal{A}$ is not $O((\log T)^c)$, for any $c > 0$; if $\mathcal{A}$ is uniformly good, this instance is either a three-armed instance or its two-armed sub-instance. Consequently no algorithm in $\mathcal{L}_{\mathrm{Async}}$ is uniformly optimal over all instances.

\paragraph{$\boldsymbol{\mathcal{L}_{\mathrm{Async}}}$ Algorithms can achieve sub-polynomial regret.}
Running the argument of the impossibility proof with general rates, we show that the rates of any asynchronous local policy must be closed under self-composition, up to constant factors, which no polylogarithmic family is. We construct closed families through a generator $\psi$, and an explicit policy whose regret lies in a sub-polynomial, super-polylogarithmic regime.


\paragraph{$\boldsymbol{\mathcal{L}_{\mathrm{Sync}}}$ Algorithms can be optimal under \emph{locally determined} randomness budgets.}
While Thompson Sampling escapes our impossibility result through synchronous index updates, standard analyses implicitly assume an unbounded source of randomness (sampling from continuous Beta distributions). In Section~\ref{sec:semilocal}, we investigate Thompson Sampling under the constraint of finite randomness, and ask how many random bits per arm per timestep are sufficient to retain asymptotic optimality. We establish that a locally determined bit budget suffices for Thompson Sampling to achieve $O(\log T)$ regret.

\paragraph{A Unified View.}
An arm whose index is frozen while it waits cannot tell how long it has waited, yet optimal exploration
revisits a suboptimal arm at times exponentially spaced in its own pull count. Some other mechanism must
supply that timing: the global clock, as in KL-UCB, or a fresh random draw for every arm at every round, as
in Thompson Sampling. Our results show that without either, optimal regret is out of reach: every optimal
index policy uses global information, either to compute its indices or to decide when to recompute them.

\section{Related Work}

Multi-armed bandits have been studied extensively \citep{thompson1933,robbins1952}, with foundational results on regret minimization established by \citet{lai1985asymptotically}, who derived the first asymptotic lower bounds for uniformly good policies. These results were later extended by \citet{burnetas1996optimal} to general parametric reward distributions, providing a more general information-theoretic characterization of optimal regret. Beyond the classical i.i.d. setting, \citet{graves1997asymptotically} developed a framework for deriving lower bounds in structured settings by analyzing controlled Markov processes, highlighting the way lower bounds change with extra prior information. 

A large body of work has focused on designing algorithms that achieve optimal regret in the general stochastic bandit setting. Prominent examples include index-based policies such as UCB \citep{auer2002finite}, KL-UCB \citep{garivier2011kl}, and IMED \citep{honda2015imed}, as well as randomized approaches such as Thompson Sampling \citep{thompson1933,agrawal2012analysis}. These algorithms achieve logarithmic regret and are asymptotically optimal in the growth rate. UCB and KL-UCB are deterministic algorithms whose index computations explicitly depend on the global timestep $t$. IMED is also deterministic, and its index requires global information in the form of the empirically best arm. Thompson Sampling, in contrast, is randomized and constructs belief distributions using only an arm's local reward history; however, it synchronously updates all arm indices at every timestep. \citet{gittins1979bandit} showed that the geometrically discounted Bayesian bandit admits an optimal index policy whose index depends only on the arm's own posterior state and the constant discount factor, and is frozen while the arm is idle; it therefore lies in $\mathcal{L}_{\mathrm{Async}}$ (with $\rho \equiv 0$). Its optimality, however, is for a stationary discounted Bayes objective: for undiscounted frequentist regret it suffers incomplete learning, and horizon-aware variants (finite-horizon Gittins; \citealt{lattimore2016finite}) read the remaining horizon, i.e.\ the global clock.

Building on the classical regret minimization framework, several works have investigated lower bounds under additional structural assumptions on the arms. For instance, \citet{combes2014unimodal} consider unimodal bandits, where the mean rewards follow a unimodal structure over an ordered set of arms, and show that this structure reduces the effective exploration complexity. Similarly, \citet{magureanu2014lipschitz} study Lipschitz bandits with finitely many arms, where smoothness constraints across arms allow for tighter regret lower bounds by exploiting similarities between nearby arms. These works demonstrate that incorporating structure across arms can significantly improve achievable regret. Note that these settings have some constraints or prior information regarding the \emph{instance}, while we consider the setting where the \emph{algorithm} is constrained to use only local arm information.

Another line of work studies decentralized and distributed multi-armed bandits. Predominantly, it is assumed that multiple agents interact with the same bandit arms, and the agents may or may not communicate. \citet{martinez2019decentralized} studied the setting with agents which can send messages to their neighbouring agents and proposed a lower bound in this setting. \citet{liu2010distributed} studied multi-agent multi-armed bandits with no communication between agents except action collisions. \citet{wang2024asynchronous} studied multi-agent settings under asynchronous agents, assuming that only a subset of agents can communicate at a time.
These settings work with different {\it agents} which can act locally, while we analyse a single central agent, with the {\it arms} acting locally.

\citet{gajane2018corrupt} and \citet{ren2020bandits} studied bandits with local differential privacy, where the arms do not show the rewards directly to the algorithm, but rather use a proxy, which can be the reward corrupted with noise. This notion of locality is different from ours, as the algorithms proposed to preserve privacy use the global timestep for index calculations, which we consider global information.

Few works have studied the limitations of restricted information classes of algorithms. In particular, purely greedy and explore-then-commit strategies are known to suffer linear regret in general, establishing $\Omega(T)$ lower bounds for such policies. \citet{kumar2021robust} studied the statistical robustness of bandit algorithms for sub-gaussian rewards with unknown variance parameters. They show that logarithmic regret is impossible, however, one can get arbitrarily close to it. These results demonstrate that restricting the adaptivity or information usage of the algorithm can fundamentally degrade performance.

Not every bandit algorithm is index-based. DMED~\citep{honda2010dmed} selects arms from a maintained
list rather than by comparing per-arm quantities, and sub-sampling duelling methods such as BESA
\citep{baransi2014besa}, SSMC~\citep{chan2020subsample}, SDA~\citep{baudry2020sda} and
LB-SDA~\citep{baudry2021lbsda} compare a leader against each challenger pairwise, so no arm carries a single
scalar value. Both families are asymptotically optimal and use cross-arm information.

We study the setting of single-agent multi-armed bandits with locally determined indices and asynchronous index updates. To the best of our knowledge, this setting has not been studied before, and it shows that global information or synchrony is necessary for logarithmic regret.

\section{Asynchronous Local Algorithms}
\label{sec:limits}

In this section we study local index algorithms with asynchronous updates. The key
observation is that if an arm is not pulled, its index is \emph{stuck}: it is a
function of that arm's own pull count and reward history, neither of which advances
while the arm is idle. Consequently an arm cannot perceive the passage of global
time, and the sequence of rounds in which a \emph{subset} of arms competes among
themselves is indistinguishable, from the point of view of those arms, from a
bandit instance consisting of that subset alone. Section~\ref{sec:embedding} makes
this precise and derives an impossibility result from it;
Section~\ref{sec:regret_families} determines what regret rates remain achievable.

\subsection{The Pathwise Embedding}
\label{sec:embedding}

\paragraph{Arms and labels.}
Throughout we distinguish an \emph{arm} --- the physical object carrying a reward
distribution and its tapes --- from its \emph{label}, the name the algorithm uses to
address it. An instance is a triple $\nu = (B, (\nu_b)_{b \in B}, \ell)$ with $B$ a
finite arm set, $\nu_b \in \mathcal{F}$, and $\ell : B \to \mathbb{N}$ an injective
labelling; we write $L := \ell(B)$. Labels are names, not tuple positions: an
instance need not use the label set $[K]$, and when we pass to a \emph{restricted
instance} $\nu|_S$ on a subset $S \subseteq B$ of arms, that instance retains both
the distributions and the labels its arms carry in $\nu$. This distinction matters
because Definition~\ref{def:async} permits the index function to depend on the
label, so an arm's index sequence is well defined only once its label is fixed;
keeping labels fixed under restriction is what allows us to compare an instance
with its sub-instance.

\paragraph{Presampled execution semantics.}
For any instance $\nu$, instantiate the environment and algorithm randomness
\emph{a priori} via infinite i.i.d.\ tapes attached to each \emph{arm} $b \in B$: a
reward tape $\mathbf{X}_b = (X_{b,1}, X_{b,2}, \dots) \sim \nu_b^{\infty}$ and a
local entropy tape $\mathbf{U}_b = (U_{b,0}, U_{b,1}, \dots)$ with
$U_{b,n} \sim \mathrm{Unif}(0,1)$. For an algorithm in
$\mathcal{L}_{\mathrm{Async}}$ the index computation after $n$ pulls requires a
random bit string $S^n_b$ of history-dependent length $\rho(L^n_b)$; since any sequence
of independent bits, finite or infinite, can be extracted deterministically from a
single uniform random variable via its binary expansion, we may take $S^n_b$ to be a
deterministic function of $U_{b,n}$.

Define the \emph{frozen index sequence} of arm $b$ by
\begin{equation}
M_b(n) \;:=\; f_{\mathrm{async}}\!\left(\ell(b),\ \big(n,\ \mathbf{X}_{b,1:n}\big),\ S^n_b\right),
\qquad n \ge 0 .
\label{eq:frozen}
\end{equation}
Thus $M_b(n)$ is a deterministic, measurable function of the label of $b$ together
with $b$'s own reward and entropy prefixes, and of nothing else: not of the global
timestep, not of the other arms, and not of which instance $b$ is placed in,
provided $b$ carries the same label. Writing
\[
\Omega(t) \;:=\; \Big\{\ell(b) \;:\; b \in \argmax_{b \in B} M_b\big(N_b(t)\big)\Big\}
\]
for the set of labels attaining the maximal index at time $t$, execution is the
deterministic recurrence
\begin{equation}
A_{t+1} \;=\; \ell^{-1}\Big(g_{\mathrm{tie}}\big(\Omega(t)\big)\Big), \qquad t \ge 0,
\label{eq:exec}
\end{equation}
started from $N_b(0) = 0$ for every $b$, so that the first selection is made from
the indices $M_b(0)$ at zero pulls.

We now show that the execution of an asynchronous local index algorithm on an
instance contains, as a subsequence, its own execution on a sub-instance. The
construction places the optimal arm at a label chosen so that the algorithm's own
tie-breaking rule works in our favour; this is what allows the lemma to hold for
\emph{every} tie-breaking rule, with no consistency requirement.

\begin{lemma}[Subsequence Pathwise Embedding]
\label{lem:subsequence_embedding}
Let $\mathcal{A} \in \mathcal{L}_{\mathrm{Async}}$ be an asynchronous local index
algorithm with index function $f_{\mathrm{async}}$ and deterministic tie-breaking rule
$g_{\mathrm{tie}}$. Let $x, y, z$ be arms with means $\mu_x > \mu_y > \mu_z$, let
$L$ be any three-element label set, and put
\[
c \;:=\; g_{\mathrm{tie}}(L), \qquad L_S \;:=\; L \setminus \{c\}.
\]
Let $\nu$ be the instance on $B = \{x,y,z\}$ with label set $L$ under any labelling
$\ell$ satisfying $\ell(x) = c$, let $S := \{y,z\}$, and let $\nu' := \nu|_S$ be the
restricted instance on $S$, carrying the labels $L_S$. Couple the tapes so that
$(\mathbf{X}_b, \mathbf{U}_b)$ is shared between $\nu$ and $\nu'$ for each $b \in S$.

For a horizon $T$, let $\tau_T := \sum_{b \in S} N^{\nu}_b(T)$ denote the number of
times an arm of $S$ is pulled in $\nu$. Then the execution of $\mathcal{A}$ on
$\nu'$ for $\tau_T$ steps reproduces exactly the subsequence of $S$-pulls in $\nu$.
Consequently, for every $b \in S$,
\begin{equation}
N^{\nu}_b(T) \;=\; N^{\nu'}_b(\tau_T).
\label{eq:embedding}
\end{equation}
\end{lemma}

\begin{proof}
Let $t_1 < t_2 < \dots < t_{\tau_T} \le T$ be the timesteps at which $\nu$ pulls an
arm of $S$, and let $A'_1, \dots, A'_{\tau_T}$ be the arms pulled by $\mathcal{A}$
on $\nu'$. We show by strong induction on $k$ that $A_{t_k} = A'_k$ as arms.

Assume $A_{t_i} = A'_i$ for all $i < k$; for $k = 1$ this hypothesis is vacuous.
Then the pull counts of the arms of $S$ agree across the two instances immediately
before the $k$-th $S$-pull:
\[
N^{\nu}_b(t_k - 1) \;=\; N^{\nu'}_b(k-1) \;=:\; n_b, \qquad b \in S.
\]
Each $b \in S$ carries the same label $\ell(b)$ in $\nu$ and in $\nu'$ and has the
same tape prefixes by the coupling, so by \eqref{eq:frozen} its index $M_b(n_b)$ is
the same number in both instances. Write $m_S := \max_{b \in S} M_b(n_b)$ for this
common maximum over $S$.

Let $\Omega := \Omega(t_k)$ denote the tied label set in $\nu$ and
$\Omega' := \Omega'(k)$ the tied label set in $\nu'$. Since $t_k$ is an $S$-pull,
the label selected in $\nu$ lies in $L_S$:
\begin{equation}
g_{\mathrm{tie}}(\Omega) \in L_S,
\qquad\text{and in particular}\qquad
\Omega \cap L_S \neq \emptyset .
\label{eq:Spull}
\end{equation}
By \eqref{eq:Spull} the maximal index in $\nu$ is attained by an arm of $S$, so that
maximum equals $m_S$; hence
\begin{equation}
\Omega' \;=\; \Omega \cap L_S ,
\label{eq:tiedsets}
\end{equation}
both sides being the set of labels of arms of $S$ whose index equals $m_S$. We
distinguish two cases according to whether the optimal arm's label $c$ is tied.

\medskip
\noindent\textbf{Case 1: $c \notin \Omega$.} Then $\Omega \subseteq L_S$, so
$\Omega = \Omega \cap L_S = \Omega'$ by \eqref{eq:tiedsets}. The aggregator
therefore evaluates the identical call $g_{\mathrm{tie}}(\Omega)$ in both instances
and selects the same label, hence the same arm.

\medskip
\noindent\textbf{Case 2: $c \in \Omega$.} We claim $|\Omega \cap L_S| = 1$. Since $|L_S| = 2$, the only alternative
is $|\Omega \cap L_S| = 2$; in that case $\Omega \supseteq \{c\} \cup L_S = L$, so
$\Omega = L$ and $g_{\mathrm{tie}}(\Omega) = g_{\mathrm{tie}}(L) = c \notin L_S$,
contradicting \eqref{eq:Spull}. Since $\Omega \cap L_S \neq \emptyset$ by
\eqref{eq:Spull}, we conclude $\Omega \cap L_S = \{\lambda\}$ for a single label
$\lambda$. Now $g_{\mathrm{tie}}(\Omega) \in \Omega \cap L_S = \{\lambda\}$, so
$\nu$ selects the arm labelled $\lambda$; and by \eqref{eq:tiedsets},
$\Omega' = \{\lambda\}$ is a singleton, so $\nu'$ selects the arm labelled $\lambda$
as well.

\medskip
In both cases $A_{t_k} = A'_k$, which completes the induction. Since the arms pulled
agree at every active step, the pull counts agree after $\tau_T$ steps of $\nu'$,
which is \eqref{eq:embedding}.
\end{proof}


\begin{remark}[Randomized tie-breaking]
\label{rem:randomized_tie}
If $g_{\mathrm{tie}}$ is randomized, Case~2 of the proof is unaffected, since it
resolves a singleton. In Case~1 the two instances evaluate $g_{\mathrm{tie}}$ on the
same set $\Omega$, so the coupling persists provided the two evaluations use the
same randomness. This holds when the tie-breaking randomness is fixed before the
interaction begins, and more generally when it is drawn from a tape indexed by the
pull counts $\big(N_b(t)\big)_{b}$ of the tied arms, since those counts agree across
the instances by the induction hypothesis. It fails only when the randomness is
indexed by the global timestep $t$, which differs between the two executions
($t_k$ in $\nu$ versus $k$ in $\nu'$). Such a rule consumes global timing
information and therefore lies outside the notion of locality studied here.
\end{remark}

Lemma~\ref{lem:subsequence_embedding} sets us up to prove the main theorem. The idea
is that a three-armed instance embeds a two-armed sub-instance over its suboptimal
arms. An algorithm with polylogarithmic regret must behave well on both. On the full
instance the suboptimal arms $y$ and $z$ are pulled polylogarithmically often, so
their combined pull count $\tau_T$ is polylogarithmic in $T$; but
Lemma~\ref{lem:subsequence_embedding} identifies $\tau_T$ as the horizon of the
sub-instance, on which $y$ is optimal and $z$ is pulled only polylogarithmically
often in $\tau_T$, that is, polylogarithmically in $\log T$. Thus the arm $z$ is
pulled far too rarely in $\nu$ to meet the Lai--Robbins lower bound there, and we
obtain a contradiction.

\begin{theorem}
\label{thm:impossibility}
Let $\mathcal{A} \in \mathcal{L}_{\mathrm{Async}}$ have an arbitrary index function
and an arbitrary deterministic tie-breaking rule. Then there exists an instance $\nu$ such that, for every $c > 0$,
\[
R_T(\mathcal{A},\nu)\ \text{is not}\ O\big((\log T)^c\big) .
\]
In particular, 
no algorithm $\mathcal{A} \in \mathcal{L}_{\mathrm{Async}}$ achieves $O(\log T)$ regret on all instances.
\end{theorem}

\begin{proof}
If $\mathcal{A}$ is not uniformly good, there are an instance $\nu$ and $\beta > 0$
such that $R_T(\mathcal{A},\nu)$ is not $o(T^\beta)$. Since
$(\log T)^c = o(T^\beta)$, this regret is not $O\big((\log T)^c\big)$ for any
$c > 0$, and the theorem holds. Assume from now on that $\mathcal{A}$ is uniformly
good.

Fix a one-parameter family $\{f(\cdot;\theta)\}_{\theta\in\Theta}\subset\mathcal{F}$
satisfying conditions (1.6), (1.7) and (1.9) of \citet{lai1985asymptotically},
take arms $x, y, z$ from it with means $\mu_x > \mu_y > \mu_z$, and let $\nu$, on the
arm set $\{x,y,z\}$, and $\nu' = \nu|_{\{y,z\}}$ be the pair of instances constructed
in Lemma~\ref{lem:subsequence_embedding}. Note that $x$ is optimal in $\nu$, that $y$
is optimal in $\nu'$, and that $z$ is suboptimal in both.

We show that the regrets on $\nu$ and $\nu'$ cannot both be polylogarithmic. Suppose,
for contradiction, that there is $\alpha > 0$ with
$R_T(\mathcal{A},\nu) = O\big((\log T)^{\alpha}\big)$ and
$R_T(\mathcal{A},\nu') = O\big((\log T)^{\alpha}\big)$. Since the suboptimality gaps
of $\nu$ and $\nu'$ are fixed positive constants and $R_T = \sum_{b \neq b^\ast} \Delta_b\, \mathbb{E}[N_b(T)]$, there exist
constants $c_1, c_2, c_3, T_0 > 0$ such that for all $t > T_0$,
\[
\mathbb{E}\big[N^{\nu}_y(t)\big] \le c_1 (\log t)^{\alpha}, \qquad
\mathbb{E}\big[N^{\nu}_z(t)\big] \le c_2 (\log t)^{\alpha}, \qquad
\mathbb{E}\big[N^{\nu'}_z(t)\big] \le c_3 (\log t)^{\alpha}.
\]
Write $\tau_T = N^{\nu}_y(T) + N^{\nu}_z(T)$, so that
$\mathbb{E}[\tau_T] \le (c_1 + c_2)(\log T)^{\alpha}$. By
Lemma~\ref{lem:subsequence_embedding} applied to the arm $z$, the identity
$N^{\nu}_z(T) = N^{\nu'}_z(\tau_T)$ holds pathwise. Under this coupling the execution on $\nu'$ is a run of $\mathcal{A}$ on $\nu'$
driven by the tapes of $y$ and $z$, so for every deterministic $t$ the random variable
$N^{\nu'}_z(t)$ has its law under $\mathbb{P}_{\nu'}$. Fix $k > 0$ and set
$\bar\tau_T := \big\lceil(\log T)^{2\alpha}\big\rceil$. Since $t \mapsto N^{\nu'}_z(t)$ is non-decreasing,
\begin{align*}
\mathbb{P}\big(N^{\nu}_z(T) \ge k \log T\big)
&= \mathbb{P}\big(N^{\nu'}_z(\tau_T) \ge k \log T\big) \\
&\le \mathbb{P}\big(N^{\nu'}_z(\tau_T) \ge k \log T,\ \tau_T < \bar\tau_T\big)
   + \mathbb{P}\big(\tau_T \ge \bar\tau_T\big) \\
&\le \mathbb{P}\big(N^{\nu'}_z(\bar\tau_T) \ge k \log T\big)
   + \mathbb{P}\big(\tau_T \ge \bar\tau_T\big).
\end{align*}
For $T$ large enough that $T > T_0$ and $\bar\tau_T > T_0$, Markov's inequality applied to
each term yields
\begin{equation}
\mathbb{P}\big(N^{\nu}_z(T) \ge k \log T\big)
\;\le\;
\frac{c_3\big(\log \bar\tau_T\big)^{\alpha}}{k \log T}
\;+\;
\frac{(c_1 + c_2)(\log T)^{\alpha}}{(\log T)^{2\alpha}} .
\label{eq:twoterms}
\end{equation}
Since $\log \bar\tau_T = O(\log\log T)$, both terms on the right-hand side vanish as
$T \to \infty$, so
\begin{equation}
\lim_{T \to \infty} \mathbb{P}\big(N^{\nu}_z(T) \ge k \log T\big) \;=\; 0
\qquad \text{for every } k > 0 .
\label{eq:vanishes}
\end{equation}
On the other hand $\mathcal{A}$ is uniformly good, so in particular condition (2.2)
of \citet{lai1985asymptotically} holds on the family $\Theta$. Their Theorem~2 is
stated for non-randomized rules, but its change-of-measure argument applies unchanged
to $\mathcal{A}$: the entropy tapes are independent of the rewards and have the same
law under both measures compared in their (2.6), so the likelihood ratio there is
unaffected. Applied to $\nu$, in which $z$ is suboptimal and $x$ is optimal, it gives,
with $D_{\nu,z} := I(\theta_z,\theta_x) \in (0,\infty)$ by their (1.6),
\[
\lim_{T \to \infty}
\mathbb{P}\!\left(N^{\nu}_z(T) \ge \frac{1-\epsilon}{D_{\nu,z}} \log T\right)
= 1 \qquad \forall\, \epsilon > 0 .
\]
Choose $k < 1/D_{\nu,z}$, and then $\epsilon > 0$ small enough that
$k \le (1-\epsilon)/D_{\nu,z}$. The display above forces
$\mathbb{P}\big(N^{\nu}_z(T) \ge k\log T\big) \to 1$, contradicting
\eqref{eq:vanishes}.

Hence, for every $\alpha > 0$, at least one of $R_T(\mathcal{A},\nu)$ and
 $R_T(\mathcal{A},\nu')$ is not $O\big((\log T)^{\alpha}\big)$.
If $R_T(\mathcal{A},\nu) = O\big((\log T)^{\alpha_1}\big)$ for some $\alpha_1 > 0$, then
$R_T(\mathcal{A},\nu')$ is not $O\big((\log T)^{\alpha}\big)$ for any
$\alpha \ge \alpha_1$, and therefore for any $\alpha > 0$, since
$O\big((\log T)^{\alpha}\big) \subseteq O\big((\log T)^{\alpha_1}\big)$ when
$\alpha \le \alpha_1$; so $\nu'$ is the required instance. Otherwise
$R_T(\mathcal{A},\nu)$ is not $O\big((\log T)^{c}\big)$ for any $c > 0$, and $\nu$ is
the required instance.
\end{proof}

\subsection{Achievable Regret Families}
\label{sec:regret_families}


Theorem~\ref{thm:impossibility} rules out polylogarithmic regret but not sublinear
regret. To see which rates remain possible, we run the argument of its proof with
general rates in place of $(\log T)^\alpha$. That proof splits the event
$\{N^{\nu}_z(T)\ge k\log T\}$ according to whether $\tau_T$ is below or above a
threshold $\bar\tau_T$, and bounds both pieces by Markov's inequality. Choosing $\bar\tau_T$
proportional to the rate of $y$ in $\nu$, the same two bounds relate the rate of $z$ in
$\nu$ to the composition of the rate of $z$ in $\nu'$ with the rate of $y$ in $\nu$.

Throughout, instances have reward distributions in $\mathcal{F}$, that is, supported on
$[0,1]$, and all hypotheses below are imposed on every such instance.

\paragraph{Rates.}
A \emph{rate} is a non-decreasing function $\phi:[1,\infty)\to[1,\infty)$ with
$\phi(t)\to\infty$; we read $\phi(T)$ as the order of the number of pulls of an arm by time
$T$. We describe a policy's pull counts by a \emph{rate family}
$\Phi=\{\phi_\gamma\}_{\gamma\in\Gamma}$, a collection of rates indexed by a set
$\Gamma$ of \emph{difficulty parameters}. The index set $\Gamma$ is arbitrary: its only
role is to name the rate that governs a given arm on a given instance, through the
assignment $\gamma_a(\nu)$ of Definition~\ref{def:realize}, and no structure on
$\Gamma$ is needed. For instance, the logarithmic rates of the Lai--Robbins bound
form the family $\{\gamma\log t\}_{\gamma\in(0,\infty)}$, with $\gamma_a(\nu)$ the
reciprocal of the constant $D_{\nu,a}$ in that bound. A rate $\phi$ is \emph{dilation-stable} if for every
$\lambda>0$ there are constants $0<d_\lambda\le D_\lambda$ with
$d_\lambda\phi(s)\le\phi(\lambda s)\le D_\lambda\phi(s)$ for all large $s$. Positive powers of
$\log t$ and of $t$, and $\exp\big((\log t)^{\gamma}\big)$ for $\gamma\in(0,1)$, are
dilation-stable; since only large arguments matter, we specify such functions by
formulas valid for large $t$.

\begin{definition}[Realizing a rate family]
\label{def:realize}
A policy $\mathcal{A}$ \emph{realizes} a family $\Phi$ of dilation-stable functions if
for every instance $\nu$ and every suboptimal arm $a$ there are
$\gamma_a(\nu)\in\Gamma$ and constants $0<c\le C$ such that
\[
\mathbb{E}_\nu\big[N_a(t)\big]\le C\,\phi_{\gamma_a(\nu)}(t)\ \ \text{for all }t\ge1,
\qquad
\lim_{T\to\infty}\mathbb{P}_\nu\big(N_a(T)\ge c\,\phi_{\gamma_a(\nu)}(T)\big)=1,
\]
and if $\phi_{\gamma_a(\nu)}=O\big(\phi_{\gamma_b(\nu)}\big)$ whenever
$\mu_a<\mu_b<\mu^\ast$, so that an arm closer to optimal is not pulled at a smaller rate.
\end{definition}

The first condition is an upper bound on the expected number of pulls. The second is a
lower bound in probability, of the kind that \citet{lai1985asymptotically} establish at
rate $\log T$ for every uniformly good policy; here it is assumed at the policy's own
rate.

\begin{theorem}[Rates compose]
\label{thm:compose}
Let $\mathcal{A}\in\mathcal{L}_{\mathrm{Async}}$ realize $\Phi$, and let $\nu$ and $\nu'$ be
the instances of Lemma~\ref{lem:subsequence_embedding}. Write
$\phi_z:=\phi_{\gamma_z(\nu)}$, $\phi_y:=\phi_{\gamma_y(\nu)}$ and
$\phi':=\phi_{\gamma_z(\nu')}$. Then there are constants $0<\kappa_1\le\kappa_2$ such that,
for all sufficiently large $T$,
\[
\kappa_1\,\phi'\big(\phi_y(T)\big)\;\le\;\phi_z(T)\;\le\;\kappa_2\,\phi'\big(\phi_y(T)\big).
\]
\end{theorem}

\begin{proof}
Let $(c_y,C_y)$, $(c_z,C_z)$ and $(c',C')$ be the constants of
Definition~\ref{def:realize} for $y$ and $z$ in $\nu$ and for $z$ in $\nu'$. By the
ordering condition there is $K$ with $\phi_z\le K\phi_y$ for large $T$. As in the proof of
Theorem~\ref{thm:impossibility}, write $\tau_T=N^{\nu}_y(T)+N^{\nu}_z(T)$, so that
$N^{\nu}_z(T)=N^{\nu'}_z(\tau_T)$ pathwise by Lemma~\ref{lem:subsequence_embedding}, and
\[
\mathbb{E}[\tau_T]\;\le\;C_y\phi_y(T)+C_z\phi_z(T)\;\le\;(C_y+C_zK)\,\phi_y(T)
\]
for large $T$. Under the coupling, the execution on $\nu'$ is a run of $\mathcal{A}$ on
$\nu'$ driven by the tapes of $y$ and $z$, so for every deterministic $t$ the random
variable $N^{\nu'}_z(t)$ has its law under $\mathbb{P}_{\nu'}$.

\emph{Upper bound.} Set $\bar\tau_T:=\big\lceil 2(C_y+C_zK)\,\phi_y(T)\big\rceil$. Splitting on
$\{\tau_T<\bar\tau_T\}$ exactly as in the proof of Theorem~\ref{thm:impossibility}, and using
that $t\mapsto N^{\nu'}_z(t)$ is non-decreasing,
\begin{align*}
\mathbb{P}\big(N^{\nu}_z(T)\ge c_z\phi_z(T)\big)
&\le\mathbb{P}\big(N^{\nu'}_z(\bar\tau_T)\ge c_z\phi_z(T)\big)+\mathbb{P}\big(\tau_T\ge \bar\tau_T\big)\\
&\le\frac{C'\phi'(\bar\tau_T)}{c_z\,\phi_z(T)}+\frac12 ,
\end{align*}
by Markov's inequality applied to each term. The left-hand side tends to $1$, so it
exceeds $3/4$ for large $T$, which gives $\phi_z(T)\le(4C'/c_z)\,\phi'(\bar\tau_T)$. Since
$\phi_y\ge1$, we have $\bar\tau_T\le\lambda\,\phi_y(T)$ with $\lambda:=2(C_y+C_zK)+1$; as $\phi'$
is non-decreasing and dilation-stable, $\phi'(\bar\tau_T)\le D_\lambda\,\phi'\big(\phi_y(T)\big)$
for large $T$. This proves the upper bound with $\kappa_2:=4C'D_\lambda/c_z$.

\emph{Lower bound.} Put $n_T:=\lfloor c_y\phi_y(T)\rfloor$, which tends to $\infty$, and
consider the event
\[
E_T:=\big\{N^{\nu}_y(T)\ge c_y\phi_y(T)\big\}\cap\big\{N^{\nu'}_z(n_T)\ge c'\phi'(n_T)\big\}.
\]
Both events have probability tending to $1$, by the lower bounds for $y$ in $\nu$ and for
$z$ in $\nu'$, the latter at the deterministic times $n_T\to\infty$; hence
$\mathbb{P}(E_T)\ge\frac12$ for large $T$. On $E_T$ we have
$\tau_T\ge N^{\nu}_y(T)\ge n_T$, so by Lemma~\ref{lem:subsequence_embedding} and
monotonicity $N^{\nu}_z(T)=N^{\nu'}_z(\tau_T)\ge N^{\nu'}_z(n_T)\ge c'\phi'(n_T)$. Taking
expectations,
\[
C_z\,\phi_z(T)\;\ge\;\mathbb{E}\big[N^{\nu}_z(T)\big]\;\ge\;\tfrac{c'}{2}\,\phi'(n_T).
\]
For large $T$ we have $n_T\ge\frac{c_y}{2}\phi_y(T)$, so
$\phi'(n_T)\ge\phi'\big(\frac{c_y}{2}\phi_y(T)\big)\ge d_{c_y/2}\,\phi'\big(\phi_y(T)\big)$.
This proves the lower bound with $\kappa_1:=c'd_{c_y/2}/(2C_z)$.
\end{proof}

Theorem~\ref{thm:compose} needs no concentration of $\tau_T$: the upper bound uses only its
mean, and the lower bound only a lower bound on it in probability. Up to constant factors,
the rate of $z$ in $\nu$ is the composite of the rate of $z$ in $\nu'$ with the rate of $y$
in $\nu$, so a family realized by a policy in $\mathcal{L}_{\mathrm{Async}}$ must be closed
under these compositions. Polylogarithmic rates fail this: if $\phi_\gamma(t)=\Theta\big((\log t)^p\big)$
for every $\gamma$, then $\phi'(\phi_y(T))=\Theta\big((\log\log T)^p\big)$ while
$\phi_z(T)=\Theta\big((\log T)^p\big)$. Polynomial rates pass it: for
$\phi_\gamma(t)=t^\gamma$ with $\gamma\in(0,1)$,
\[
\phi_a\big(\phi_b(t)\big)\;=\;t^{ab}\;=\;\phi_{ab}(t).
\]
More generally, let $\psi$ be a continuous, strictly increasing bijection from an interval
$(t_0,\infty)$ onto $(0,\infty)$. It \emph{generates} the family
\begin{equation}
\phi_\gamma(t) \;:=\; \psi^{-1}\!\big(\gamma\,\psi(t)\big), \qquad \gamma\in(0,1),
\label{eq:generator}
\end{equation}
which is closed under composition, since
$\phi_a\circ\phi_b=\psi^{-1}\big(ab\,\psi(t)\big)=\phi_{ab}$; the generator $\psi=\log$
gives the polynomial family. We now construct an asynchronous local policy whose pull counts
are bounded by the family generated by a given $\psi$. With $\psi=\log$ its regret is
polynomial, and slower generators give sub-polynomial regret.

Two structural features guide the design of such an index.
First, \textbf{decay with pulls}: an unpulled arm's index is frozen, so the arm
currently being pulled must eventually decrease its own index, or no other arm can
ever be selected; we therefore require the index to be decreasing in the pull count
for a fixed empirical mean. Second, \textbf{shared infimum}: all arms must share a
common infimum of index values, or decay to $-\infty$, since otherwise an arm whose
index is bounded below by the supremum of another arm's range can permanently
dominate the optimal arm, incurring linear regret.

\begin{definition}[Local $\psi$-index]
\label{def:psi_index}
Let $\psi : \mathbb{R}_{>0} \to \mathbb{R}_{>0}$ be monotonically increasing. The
\emph{local $\psi$-index} of an arm with $n$ pulls and empirical mean $u$ is
\[
I_\psi(u,n) \;=\; -(1-u)\,\psi(n).
\]
\end{definition}

Both features hold: for fixed $u < 1$ the index is strictly decreasing in $n$ and
tends to $-\infty$. We now show that the pull counts of the resulting policy are
bounded above by the rate family \eqref{eq:generator} generated by $\psi$.

\begin{theorem}
\label{thm:psi_index}
Let $\psi$ be monotonically increasing and such that the generated family
$\phi_\gamma(t) := \psi^{-1}\big(\gamma\psi(t)\big)$ satisfies
$\phi_\gamma(t) = \omega(\log t)$ for every $\gamma \in (0,1)$. Consider the
asynchronous local index algorithm given by the index $I_\psi$, run on an instance
$\nu$ with optimal arm $b^\ast$ of mean $\mu^\ast$. Then for every suboptimal arm
$b$ with mean $\mu_b < \mu^\ast$, and every $\epsilon > 0$ small enough that
$\mu_b + \epsilon < \mu^\ast - \epsilon$,
\[
\mathbb{E}\big[N_b(T)\big] \;\le\; C(\epsilon) \;+\; \phi_{\gamma(\epsilon)}(T),
\qquad
\gamma(\epsilon) := \frac{1-\mu^\ast+\epsilon}{1-\mu_b-\epsilon} \;<\; 1,
\]
where $C(\epsilon)$ is finite and does not depend on $T$.
\end{theorem}

Taking $\psi(t) = \log^{(m)} t$, the $m$-fold iterated logarithm, gives
$\phi_\gamma(t) = \exp^{(m-1)}\!\big((\log^{(m-1)} t)^{\gamma}\big)$ and hence
\[
\mathbb{E}\big[N_b(T)\big] \;\le\;
\exp^{(m-1)}\!\left(\big(\log^{(m-1)} T\big)^{\frac{1-\mu^\ast+\epsilon}{1-\mu_b-\epsilon}}\right)
+ C(\epsilon),
\]
which is super-polylogarithmic for every $m \ge 1$ and, for $m \ge 2$,
sub-polynomial; each of these families is closed, being of the form
\eqref{eq:generator}. Together with
Theorem~\ref{thm:impossibility} this places the achievable regret of
$\mathcal{L}_{\mathrm{Async}}$ strictly between the polylogarithmic and polynomial
regimes. The appendix reports experiments with several generators, confirming the
bound of Theorem~\ref{thm:psi_index}.

\begin{proof}
Write $\gamma := \gamma(\epsilon)$ and $\ell := \phi_\gamma(T) = \psi^{-1}(\gamma\psi(T))$.
Since at most $\lceil \ell \rceil$ pulls of $b$ can occur while $N_b(t-1) < \ell$,
\[
\mathbb{E}\big[N_b(T)\big]
\;\le\; \ell + 1 + \sum_{t=1}^{T} \mathbb{P}\big(A_t = b,\ N_b(t-1) \ge \ell\big).
\]
Abbreviate $\hat\mu_b := \hat\mu_{b,N_b(t-1)}$ and
$\hat\mu^\ast := \hat\mu_{b^\ast,N_{b^\ast}(t-1)}$. The event
$\{A_t = b,\ N_b(t-1) \ge \ell\}$ is contained in
$E_{1,t} \cup E_{2,t} \cup E_{3,t}$, where
\begin{align*}
E_{1,t} &:= \big\{A_t = b,\ N_b(t-1) \ge \ell,\ \hat\mu_b \le \mu_b + \epsilon,\
  \hat\mu^\ast \ge \mu^\ast - \epsilon \big\}, \\
E_{2,t} &:= \big\{A_t = b,\ \hat\mu_b > \mu_b + \epsilon \big\}, \\
E_{3,t} &:= \big\{A_t = b,\ N_b(t-1) \ge \ell,\ \hat\mu_b \le \mu_b + \epsilon,\
  \hat\mu^\ast < \mu^\ast - \epsilon \big\} :
\end{align*}
indeed, on $\{A_t = b,\ N_b(t-1)\ge\ell\}$ either $\hat\mu_b > \mu_b + \epsilon$,
which is $E_{2,t}$, or $\hat\mu_b \le \mu_b + \epsilon$, in which case $E_{1,t}$ or
$E_{3,t}$ holds according to the value of $\hat\mu^\ast$. Note that $E_{1,t}$ and
$E_{3,t}$ both contain the selection event $A_t = b$, so the index comparison below
is available on each.

\textbf{1. Clean event.} On $E_{1,t}$ the arm $b$ is selected, so its index is
maximal; in particular
$I_\psi(\hat\mu_b, N_b(t-1)) \ge I_\psi(\hat\mu^\ast, N_{b^\ast}(t-1))$, that is,
\[
(1-\hat\mu_b)\,\psi\big(N_b(t-1)\big) \;\le\; (1-\hat\mu^\ast)\,\psi\big(N_{b^\ast}(t-1)\big).
\]
The defining inequalities of $E_{1,t}$ give $1-\hat\mu_b \ge 1-\mu_b-\epsilon$ and
$1-\hat\mu^\ast \le 1-\mu^\ast+\epsilon$, while $N_b(t-1) \ge \ell$ and monotonicity
of $\psi$ give $\psi(N_b(t-1)) \ge \psi(\ell) = \gamma\psi(T)$. Hence
\[
(1-\mu_b-\epsilon)\,\gamma\,\psi(T) \;\le\; (1-\mu^\ast+\epsilon)\,\psi\big(N_{b^\ast}(t-1)\big).
\]
By definition of $\gamma$, $(1-\mu_b-\epsilon)\gamma = 1-\mu^\ast+\epsilon \ge \epsilon > 0$,
so dividing and applying $\psi^{-1}$ yields $T \le N_{b^\ast}(t-1) \le t-1 < T$, a
contradiction. Thus $\mathbb{P}(E_{1,t}) = 0$ for all $t$.

\textbf{2. Overestimation of the suboptimal arm.} Since $N_b$ increases by one at each
$t$ with $A_t = b$, each value $s$ of $N_b(t-1)$ occurs on at most one such $t$, so
$\sum_{t=1}^{T}\mathbb{1}\{E_{2,t}\} \le \sum_{s\ge1}\mathbb{1}\{\hat\mu_{b,s} > \mu_b+\epsilon\}$,
where $\hat\mu_{b,s}$ is the mean of the first $s$ entries of $b$'s reward tape. Taking
expectations and applying Hoeffding's inequality to this i.i.d.\ tape,
\[
\sum_{t=1}^{T} \mathbb{P}(E_{2,t}) \;\le\; \sum_{s=1}^{\infty} e^{-2s\epsilon^2}
\;=\; \frac{1}{e^{2\epsilon^2}-1} \;=:\; C_0(\epsilon) \;<\; \infty .
\]

\textbf{3. Underestimation of the optimal arm.} On $E_{3,t}$ the same index
comparison holds; using $1-\hat\mu^\ast \le 1$ in place of the bound from $E_{1,t}$,
\[
N_{b^\ast}(t-1) \;\ge\; \psi^{-1}\!\big((1-\mu^\ast+\epsilon)\,\psi(T)\big) \;=:\; L .
\]
If $1-\mu^\ast+\epsilon \ge 1$ then $L \ge T$, impossible since
$N_{b^\ast}(t-1) < T$, so $\mathbb{P}(E_{3,t}) = 0$. Otherwise
$L = \phi_{1-\mu^\ast+\epsilon}(T) = \omega(\log T)$ by hypothesis, so for any
$\lambda$ with $2\lambda\epsilon^2 > 1$ there is $T_1(\lambda)$ with
$L \ge \lambda\log T$ for $T > T_1(\lambda)$. Bounding by a union over the values
$s \ge L$ of $N_{b^\ast}(t-1)$ and applying Hoeffding,
\[
\sum_{t=1}^{T} \mathbb{P}(E_{3,t})
\;\le\; T \sum_{s \ge L} e^{-2s\epsilon^2}
\;\le\; \frac{T\,e^{-2L\epsilon^2}}{1-e^{-2\epsilon^2}}
\;\le\; \frac{T^{\,1-2\lambda\epsilon^2}}{1-e^{-2\epsilon^2}},
\]
which is bounded in $T$ since $1-2\lambda\epsilon^2 < 0$; for $T \le T_1(\lambda)$
the sum is at most $T_1(\lambda)$. Hence
$\sum_{t=1}^{T}\mathbb{P}(E_{3,t}) \le C_1(\epsilon) < \infty$ uniformly in $T$.

\textbf{Conclusion.} Summing the three contributions gives
$\mathbb{E}[N_b(T)] \le \phi_{\gamma(\epsilon)}(T) + C(\epsilon)$ with
$C(\epsilon) := 1 + C_0(\epsilon) + C_1(\epsilon)$.
\end{proof}

\begin{remark}[Why synchronous policies escape]
\label{rem:sync_escape}
The embedding of Lemma~\ref{lem:subsequence_embedding} rests on one fact: an idle arm's index is a function
of its own pull count. A synchronous policy draws each arm's bit string afresh at every round, so its
randomness is indexed by the global time $t$ rather than by $N_a(t)$. In the coupling of
Lemma~\ref{lem:subsequence_embedding} the $k$-th $S$-pull occurs at time $t_k$ in $\nu$ and at time $k$ in
$\nu'$, so the two executions read different randomness and need not agree. This is how Thompson Sampling,
which is local but synchronous, attains $O(\log T)$ regret
\citep{agrawal2012analysis,kaufmann2012thompson}: resampling at every round lets an idle arm's index move,
at the price of every arm being told when a round occurs.
\end{remark}

\section{Synchronous Local Index Algorithms}
\label{sec:semilocal}
Thompson Sampling, unlike the algorithms in $\mathcal{L}_{\mathrm{Async}}$, samples from each arm's posterior distribution at all timesteps. Hence, it escapes the embedding lemma, and incurs logarithmic regret. However, it also samples from continuous Beta distributions, which require an infinite number of random bits. In this section, we analyze the amount of randomness which suffices for Thompson Sampling to remain optimal. 
We show that under sufficient $B(L_a^t)$, synchronous local index algorithms are optimal. 


\begin{theorem}[Local Bit Growth Sufficiency for Thompson Sampling]
\label{thm:bit_growth}
Consider a Bernoulli instance $\nu$ with arms labelled so that $\mu_1 > \mu_2 \ge \cdots \ge \mu_K$. Let $\mu_a$ denote the mean reward of arm $a$, and let $B(L_a^t) = h(N_a(t))$ be
the number of bits available to arm $a$ at timestep $t$. At each timestep $t$, arm 
$a$ draws ${U}_{a,t}$ uniformly from $\left\{k/2^{B(L_a^t)} : k = 1, \ldots, 
2^{B(L_a^t)}\right\}$ and the algorithm pulls:
\[
A_t = \argmax_{a \in [K]}\; Q_{a,t}({U}_{a,t})
\]
where $Q_{a,t} = F_{a,t}^{-1}$ is the quantile function of arm $a$'s posterior 
$\mathrm{Beta}(1 + S_a(t), 1 + N_a(t) - S_a(t))$. Let $\hat{\pi}_{TS} \in 
\mathcal{L}_{\mathrm{Sync}}$ denote this algorithm. Then, if $h(n) = \omega(n)$ and monotonically increasing:
\[
R_T(\hat{\pi}_{TS}, \nu) = O(\log T)
\]
\end{theorem}



\begin{proof}[Proof of Theorem~\ref{thm:bit_growth}]
We follow the analysis of \citet{kaufmann2012thompson}, henceforth \emph{KKM}, 
adopting their notation throughout. Couple $U_{a,t} = \lceil 2^{B_a(t)} 
V_{a,t}\rceil / 2^{B_a(t)}$ to $V_{a,t} \sim \mathrm{Uniform}[0,1)$, and let 
$\hat\theta_{a,t} = Q_{a,t}(U_{a,t})$ and $\theta_{a,t} = Q_{a,t}(V_{a,t})$ 
denote the discretized and continuous Thompson samples respectively. By 
monotonicity of $Q_{a,t}$ under the floor map:
\begin{equation}\label{eq:coupling}
\hat\theta_{a,t} \geq \theta_{a,t} \qquad \text{surely.}
\end{equation}
Also note that under this quantization scheme, for any $x \in (0,1)$,
\[
\mathbb{P}(\hat\theta_{a,t} \geq x |N_a(t)=B) \leq \mathbb{P}(\theta_{a,t} \geq x |N_a(t)=B) + \frac{1}{2^B}
\]
Now, the expected number of suboptimal draws decomposes as:
\begin{align*}
\mathbb{E}[N_{a,T}] &\leq \underbrace{\sum_{t=1}^{T} \mathbb{P}\!\left(\hat\theta_{1,t} \leq 
\mu_1 - \sqrt{\frac{6\log t}{N_{1,t}}}\right)}_{A} + 
\underbrace{\sum_{t=1}^{T} \mathbb{P}\!\left(u_{a,t} > \mu_1 - 
\sqrt{\frac{6\log t}{N_{1,t}}},\, A_t = a\right)}_{B} \\
&\quad + \underbrace{\sum_{t=1}^T \mathbb{P}(\hat\theta_{a,t}>q_{a,t}, A_t=a)}_{C}
\end{align*}
where $u_{a,t}$ and $q_{a,t}$ are the KL-UCB and Bayes-UCB indices of KKM.

\textbf{Term A and Lemma 3.} Write $\delta := \frac{\mu_1-\mu_2}{2}$ for the half-gap to the best
suboptimal arm, as in KKM's own Lemma 3. Since $\hat\theta_{a,t} \geq \theta_{a,t}$ by
\eqref{eq:coupling}, we have $\mathbb{P}(\hat\theta_{1,t} \leq \mu_1-\sqrt{\frac{6\log t}{N_{1,t}}}) \leq \mathbb{P}(\theta_{1,t} \leq \mu_1-\sqrt{\frac{6\log t}{N_{1,t}}})$, so $(A) = O(1)$ exactly as in KKM. Similarly,
\[
\mathbb{P}\!\left(\forall s \in J,\; \hat\theta_{1,s} \leq \mu_2+\delta,\; 
|J| \geq f(t)\right) \leq 
\mathbb{P}\!\left(\forall s \in J,\; \theta_{1,s} \leq \mu_2+\delta,\; 
|J| \geq f(t)\right)
\]
so Lemma 3 of KKM holds automatically under ceil quantization.

\textbf{Lemma 4.} 
The original Lemma 4 in KKM states 
\[
\mathbb{P}(\exists s\leq t, \exists a \neq 1:\theta_{a,s} > \mu_a+\delta_a, N_{a,s} > C_a\log t) \leq \frac{2(K-1)}{t^2}
\]
Under finite precision, we have:
\[
\begin{aligned}
&\mathbb{P}\!\left(\exists s\leq t,\ \exists a \neq 1 :
\hat\theta_{a,s} > \mu_a+\delta_a,\ 
N_{a,s} > C_a\log t \right) \\
&\leq (K-1)\sum_{s=1}^{t}
\mathbb{P}\!\left(
\hat\theta_{a,s} > \mu_a+\delta_a,\ 
N_{a,s} > C_a\log t
\right) \\
&\leq (K-1)\sum_{s=1}^{t}
\mathbb{P}\!\left(
\hat\theta_{a,s} > \mu_a+\delta_a
\mid
N_{a,s} > C_a\log t
\right)
\mathbb{P}\!\left(
N_{a,s}>C_a\log t
\right) \\
&\leq (K-1)\sum_{s=1}^{t}
\left[
\mathbb{P}\!\left(
\theta_{a,s} > \mu_a+\delta_a
\right)
+
\frac{1}{2^{h(C_a\log t)}}
\right] \\
&\leq \frac{2(K-1)}{t^2}
+
\frac{t}{2^{\omega(\log t)}} .
\end{aligned}
\]
The extra term is $o(\frac{1}{t^k})$, and hence, does not affect the analysis. Effectively, Lemma 4 holds.

\textbf{Term C.} 
In KKM, it is shown that
\[
\sum_{t=1}^T 
\mathbb{P}\!\left(
\theta_{a,t}>q_{a,t} 
\right) \leq2
\]
We can write 
\[
\begin{aligned}
&\sum_{t=1}^T 
\mathbb{P}\!\left(
\hat\theta_{a,t}>q_{a,t},\ A_t=a
\right) \\
&\leq 
\sum_{t=1}^T 
\mathbb{P}\!\left(
A_t=a,\ N_{a,t}\leq h^{-1}(\log_2 T)
\right)
+
\sum_{t=1}^T 
\mathbb{P}\!\left(
\hat\theta_{a,t}>q_{a,t},\ 
N_{a,t}> h^{-1}(\log_2 T)
\right) \\
&\leq 
h^{-1}(\log_2 T)
+
\sum_{t=1}^T 
\mathbb{P}\!\left(
\theta_{a,t}>q_{a,t}
\right)
+
\sum_{t=1}^T 
\frac{1}{2^{h(h^{-1}(\log_2 T)}} \\
&\leq 
h^{-1}(\log_2 T)
+2+1\\ 
& = o(\log_2 T)
\end{aligned}
\]

\textbf{Proposition 1.} KKM show that $\sum_{t=1}^{\infty} \mathbb{P}(N_{1,t} \leq 
t^b) \leq C_b < \infty$ for some $b \in (0,1)$, by bounding $P(\xi_j \geq 
t^{1-b}-1)$ via their Lemma 3 and 4. Since we show that Lemma 3 holds, and Lemma 4 also holds with an
additional finite-randomness penalty, Proposition 1 holds.

\textbf{Term B.} The analysis of Term B in KKM involves only the KL-UCB 
index $u_{a,t}$ and the arm pull counts $N_{a,t}$, neither of which depend 
on the Thompson samples $\hat\theta_{a,t}$. Hence the proof of Lemma 2 of 
KKM carries through unchanged, yielding:
\[
(B) \leq (1+\epsilon)\frac{\log T + \log\log T}{K(\mu_a,\mu_1)} + 
D(\epsilon,\mu_1,\mu_a).
\]

\textbf{Conclusion.} Combining all terms, for every $\epsilon > 0$:
\[
E[N_{a,T}] \leq (1+\epsilon)\frac{\log T + \log\log T}{K(\mu_a,\mu_1)} 
+ D(\epsilon,\mu_1,\mu_a) + O(1) + o(\log_2 T)
\]
and therefore $R_T(\hat{\pi}_{TS},\nu) = O(\log T)$.
\end{proof}

Note that this demonstrates sufficiency, not necessity. It may be possible to be optimal with a fewer number
of bits as well. We know that with 0 bits per arm, we essentially have a deterministic algorithm, and this
also resides in $\mathcal{L}_{\mathrm{Async}}$, and thus cannot incur logarithmic regret. However, by introducing just 1 bit per arm, the embedding breaks. By Theorem~\ref{thm:bit_growth}, we know that $B(L_a^t) > cN_{a,t}$ is sufficient for optimal regret. Hence, there must be a transition point and a minimum necessary bit-growth rate required for a synchronous local index algorithm to be optimal. This remains an open question, but we believe that a smartly chosen index might be able to achieve logarithmic regret with just 1 bit of randomness per arm. 

\section{Limitations and Future Work}
For $\mathcal{L}_{\mathrm{Async}}$ we show that super-polylogarithmic regret is achievable, but the minimum
achievable rate is unknown. Concretely: for any $g(n) = \omega(\log n)$, does there exist a family
$\{\phi_\gamma(t): \gamma \in (0,1)\}$ with $\phi_b(\phi_c(t)) = \phi_{bc}(t)$ for all $b,c \in (0,1)$ and
$\phi_\gamma(t) = o(g(t))$ for all $\gamma \in (0,1)$? We also designed the policies so that the
instance-dependent constants interact multiplicatively. Is there another interaction giving a rate
arbitrarily close to polylogarithmic?

We have only partially addressed how much randomness a synchronous local policy needs. A synchronous index
with no random bits changes only when its arm is pulled, so it is asynchronous and
Theorem~\ref{thm:impossibility} applies to it; Theorem~\ref{thm:bit_growth} shows that a bit budget growing
faster than linearly in the pull count already suffices for optimal regret, in place of the infinitely many
bits Thompson Sampling draws from its continuous posterior. This is a sufficiency result only: the minimum
bit-growth rate needed for optimality, and whether a single bit per arm can already suffice with a more
carefully chosen index, remain open.

\section{Concluding Remarks}

An arm whose index depends only on its own history, and is recomputed only when the arm is pulled, cannot
perceive time while it waits. We have shown that this rules out optimal regret: by
Lemma~\ref{lem:subsequence_embedding}, the suboptimal arms of a three-armed instance replay a two-armed
instance on a clock that is exponentially compressed, and polylogarithmic regret cannot hold on both.
The rates of an asynchronous local policy that realizes a rate family must be closed under
self-composition, which no polylogarithmic family is; closed families exist strictly between
polylogarithmic and polynomial, and we have exhibited policies whose pull counts are bounded by such rates.

Optimal index policies break the embedding in one of two ways. KL-UCB and UCB compute their indices from the
global timestep. Thompson Sampling computes each index from local information only, but resamples every
arm at every round, so each arm must be told \emph{when} a round occurs; this requires either a globally
synchronized clock or knowledge of when other arms are pulled, both of which are global information. In
this sense every optimal index policy uses global information, either to compute its indices or to decide
when to recompute them.


\bibliography{references}  

\newpage
\appendix

\section{Experiments}

\subsection{Experiments with Asynchronous Local Index Policies}
We compare 4 generators:
$\psi(t) = \log t$ which implies $\phi_a(t) = \psi^{-1}(a.\psi(t)) = t^a$,
$\psi(t) = \sqrt t$ which implies $\phi_a(t) = \psi^{-1}(a.\psi(t)) = a^2t$,
$\psi(t) = \log \log t$ which implies $\phi_a(t) = \psi^{-1}(a.\psi(t)) = \exp((\log t)^a)$  and
$\psi(t) = \sqrt{\log t}$ which gives $\phi_a(t) = \psi^{-1}(a.\psi(t)) = t^{a^2}$

\begin{figure}[H]
    \centering
    \includegraphics[scale=0.6]{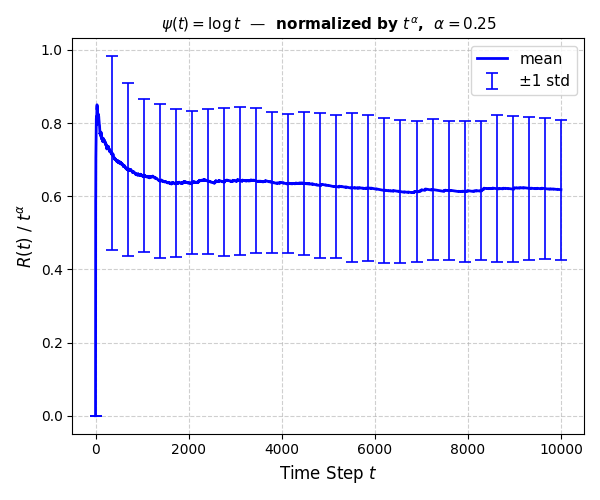}
    \caption[Normalized regret vs time plot showing $\Theta(t^{0.25})$ regret (instance (0.8,0.2), $\psi(t)=\log t$)]{The figure shows the $\frac{R(t)}{t^{0.25}}$ vs $t$ plot with the generator $\psi(t) = \log t$ on the instance (0.8,0.2). The  plot flattens with time, showing that $R(t) = \Theta(t^{0.25})$ as expected.}
    \label{fig:1}
\end{figure}

\begin{figure}[H]
    \centering
    \includegraphics[scale=0.6]{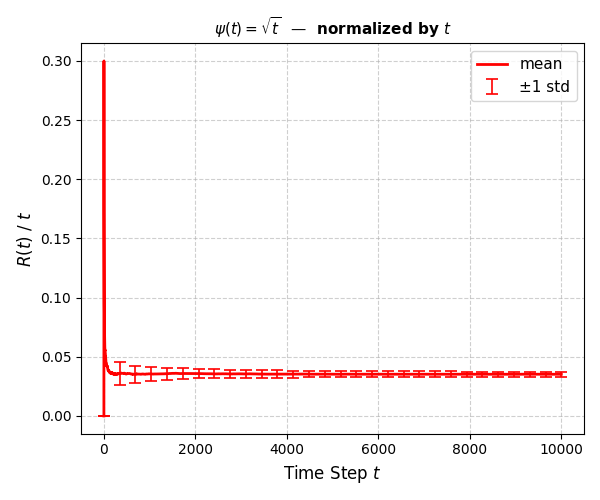}
    \caption[Normalized regret vs time plot showing $\Theta(t)$ regret (instance (0.8,0.2), $\psi(t)=\sqrt t$)]{The figure shows the $\frac{R(t)}{t}$ vs $t$ plot with the generator $\psi(t) = \sqrt t$ on the instance (0.8,0.2). The  plot flattens with time, showing that $R(t) = \Theta(t)$ as expected.}
    \label{fig:2}
\end{figure}

\begin{figure}[H]
    \centering
    \includegraphics[scale=0.6]{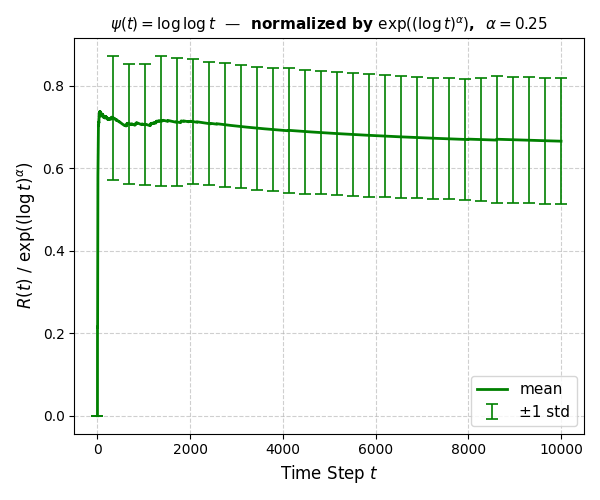}
    \caption[Normalized regret vs time plot showing $\Theta(\exp({(\log t)^{0.25}}))$ regret (instance (0.8,0.2), $\psi(t)=\log\log t$)]{The figure shows the $\frac{R(t)}{\exp({(\log t)^{0.25}})}$ vs $t$ plot with the generator $\psi(t) = \log \log t$ on the instance (0.8,0.2). The  plot flattens with time, showing that $R(t) = \Theta(\exp({(\log t)^{0.25}}))$ as expected.}
    \label{fig:3}
\end{figure}

\begin{figure}[H]
    \centering
    \includegraphics[scale=0.6]{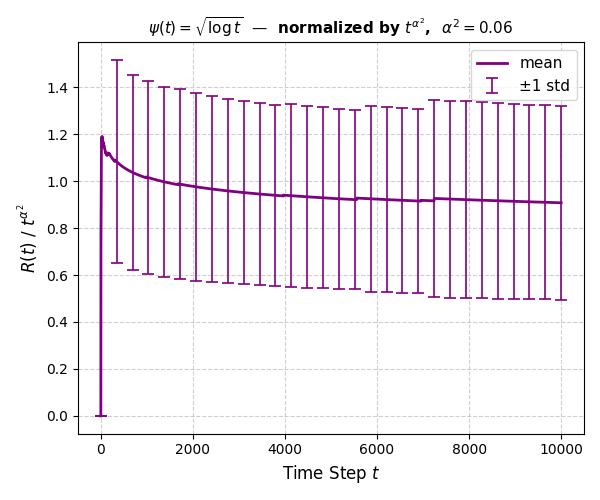}
    \caption[Normalized regret vs time plot showing $\Theta(t^{0.25^2})$ regret (instance (0.8,0.2), $\psi(t)=\sqrt{\log t}$)]{The figure shows the $\frac{R(t)}{t^{0.25^2}}$ vs $t$ plot with the generator $\psi(t) = \sqrt{\log t}$ on the instance (0.8,0.2). The  plot flattens with time, showing that $R(t) = \Theta(t^{0.25^2})$ as expected.}
    \label{fig:4}
\end{figure}

\begin{figure}[H]
    \centering
    \includegraphics[scale=0.6]{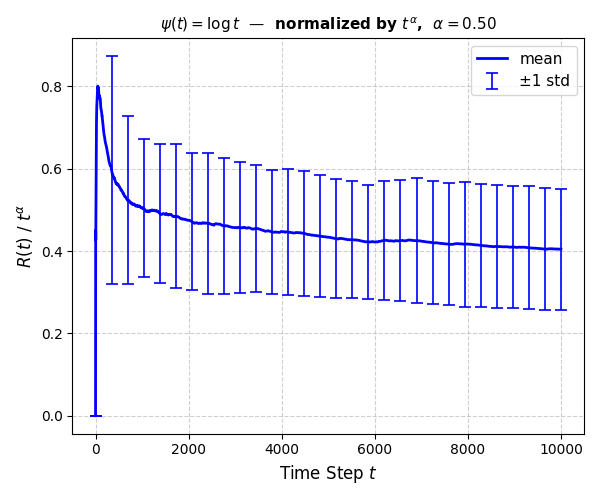}
    \caption[Normalized regret vs time plot showing $\Theta(t^{0.5})$ regret (instance (0.8,0.6,0.5,0.4), $\psi(t)=\log t$)]{The figure shows the $\frac{R(t)}{t^{0.5}}$ vs $t$ plot with the generator $\psi(t) = \log t$ on the instance (0.8,0.6,0.5,0.4). The  plot flattens with time, showing that $R(t) = \Theta(t^{0.5})$ as expected.}
    \label{fig:5}
\end{figure}

\begin{figure}[H]
    \centering
    \includegraphics[scale=0.6]{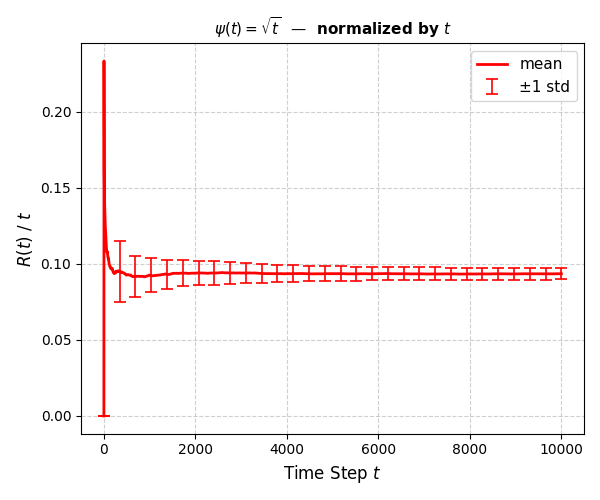}
    \caption[Normalized regret vs time plot showing $\Theta(t)$ regret (instance (0.8,0.6,0.5,0.4), $\psi(t)=\sqrt t$)]{The figure shows the $\frac{R(t)}{t}$ vs $t$ plot with the generator $\psi(t) = \sqrt t$ on the instance (0.8,0.6,0.5,0.4). The  plot flattens with time, showing that $R(t) = \Theta(t)$ as expected.}
    \label{fig:6}
\end{figure}

\begin{figure}[H]
    \centering
    \includegraphics[scale=0.6]{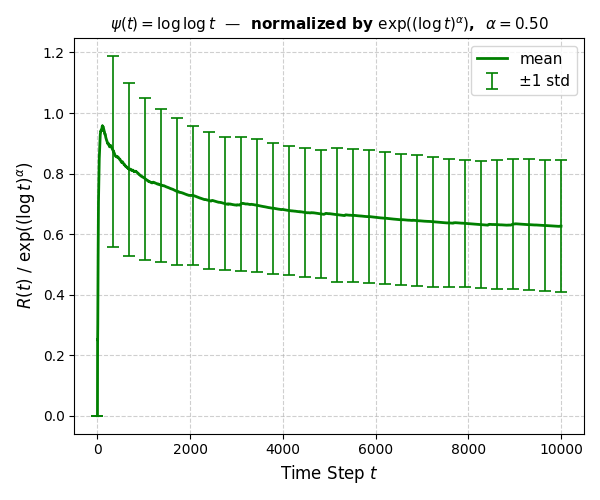}
    \caption[Normalized regret vs time plot showing $\Theta(\exp({(\log t)^{0.5}}))$ regret (instance (0.8,0.6,0.5,0.4), $\psi(t)=\log\log t$)]{The figure shows the $\frac{R(t)}{\exp({(\log t)^{0.5}})}$ vs $t$ plot with the generator $\psi(t) = \log \log t$ on the instance (0.8,0.6,0.5,0.4). The  plot flattens with time, showing that $R(t) = \Theta(\exp({(\log t)^{0.5}}))$ as expected.}
    \label{fig:7}
\end{figure}

\begin{figure}[H]
    \centering
    \includegraphics[scale=0.6]{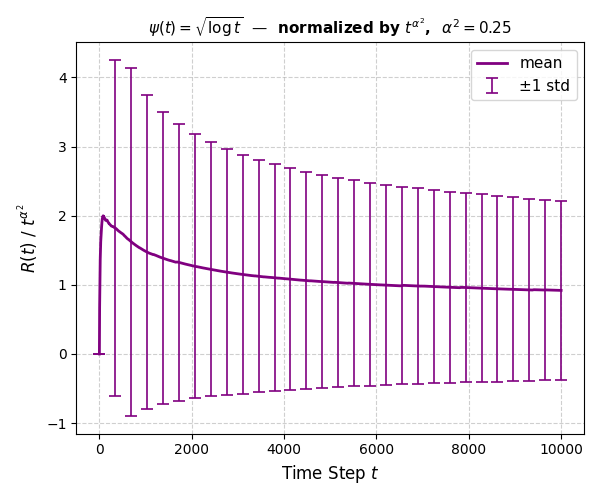}
    \caption[Normalized regret vs time plot showing $\Theta(t^{0.5^2})$ regret (instance (0.8,0.6,0.5,0.4), $\psi(t)=\sqrt{\log t}$)]{The figure shows the $\frac{R(t)}{t^{0.5^2}}$ vs $t$ plot with the generator $\psi(t) = \sqrt{\log t}$ on the instance (0.8,0.6,0.5,0.4). The  plot flattens with time, showing that $R(t) = \Theta(t^{0.5^2})$ as expected.}
    \label{fig:8}
\end{figure}

\begin{figure}[H]
    \centering
    \includegraphics[scale=0.6]{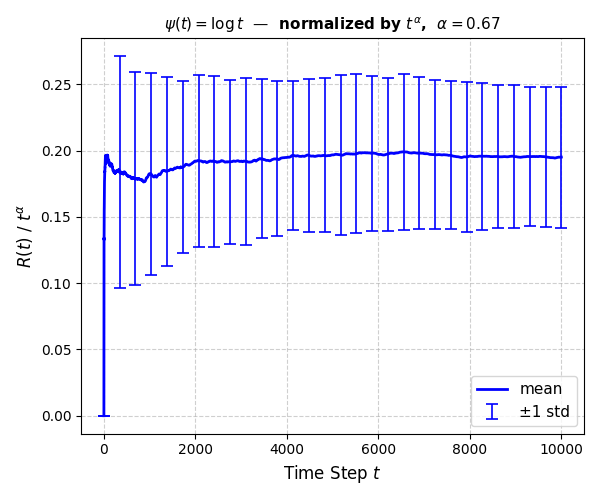}
    \caption[Normalized regret vs time plot showing $\Theta(t^{0.67})$ regret (instance (0.6,0.4), $\psi(t)=\log t$)]{The figure shows the $\frac{R(t)}{t^{0.67}}$ vs $t$ plot with the generator $\psi(t) = \log t$ on the instance (0.6,0.4). The  plot flattens with time, showing that $R(t) = \Theta(t^{0.67})$ as expected.}
    \label{fig:9}
\end{figure}

\begin{figure}[H]
    \centering
    \includegraphics[scale=0.6]{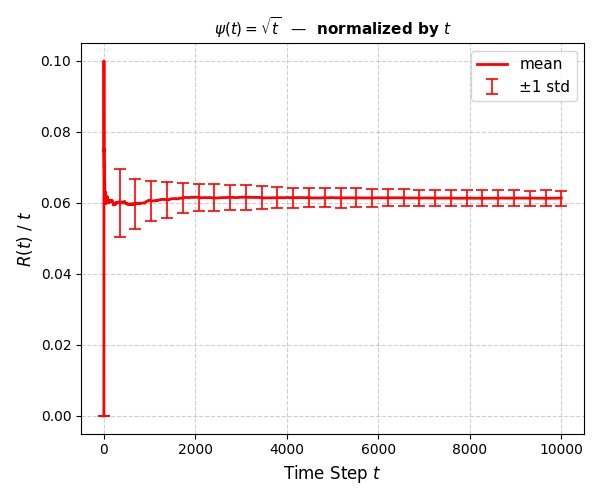}
    \caption[Normalized regret vs time plot showing $\Theta(t)$ regret (instance (0.6,0.4), $\psi(t)=\sqrt t$)]{The figure shows the $\frac{R(t)}{t}$ vs $t$ plot with the generator $\psi(t) = \sqrt t$ on the instance (0.6,0.4). The  plot flattens with time, showing that $R(t) = \Theta(t)$ as expected.}
    \label{fig:10}
\end{figure}

\begin{figure}[H]
    \centering
    \includegraphics[scale=0.6]{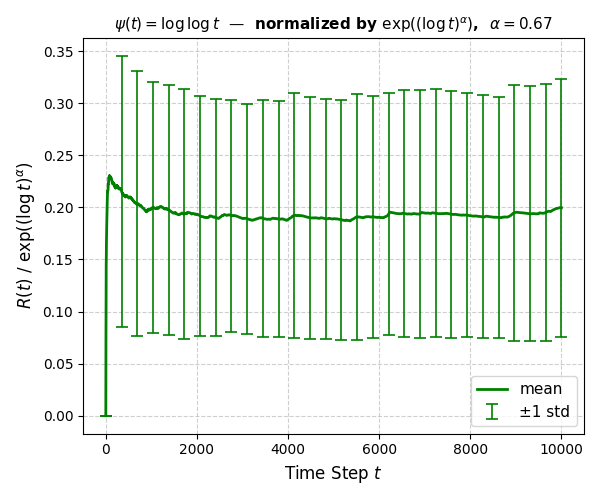}
    \caption[Normalized regret vs time plot showing $\Theta(\exp({(\log t)^{0.67}}))$ regret (instance (0.6,0.4), $\psi(t)=\log\log t$)]{The figure shows the $\frac{R(t)}{\exp({(\log t)^{0.67}})}$ vs $t$ plot with the generator $\psi(t) = \log \log t$ on the instance (0.6,0.4). The  plot flattens with time, showing that $R(t) = \Theta(\exp({(\log t)^{0.67}}))$ as expected.}
    \label{fig:11}
\end{figure}

\begin{figure}[H]
    \centering
    \includegraphics[scale=0.6]{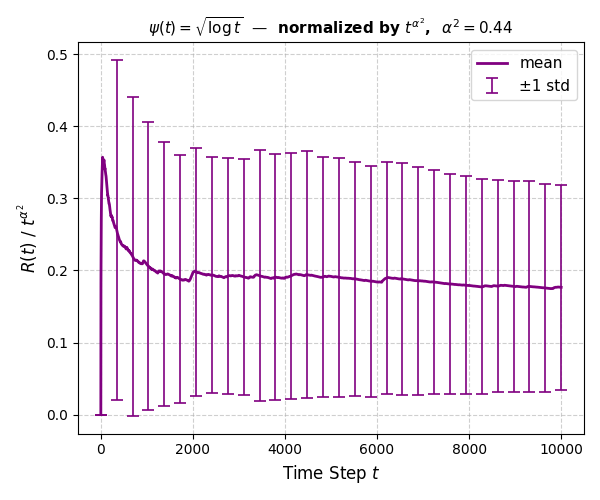}
    \caption[Normalized regret vs time plot showing $\Theta(t^{0.67^2})$ regret (instance (0.6,0.4), $\psi(t)=\sqrt{\log t}$)]{The figure shows the $\frac{R(t)}{t^{0.67^2}}$ vs $t$ plot with the generator $\psi(t) = \sqrt{\log t}$ on the instance (0.6,0.4). The  plot flattens with time, showing that $R(t) = \Theta(t^{0.67^2})$ as expected.}
    \label{fig:12}
\end{figure}

\begin{figure}[H]
    \centering
    \includegraphics[scale=0.6]{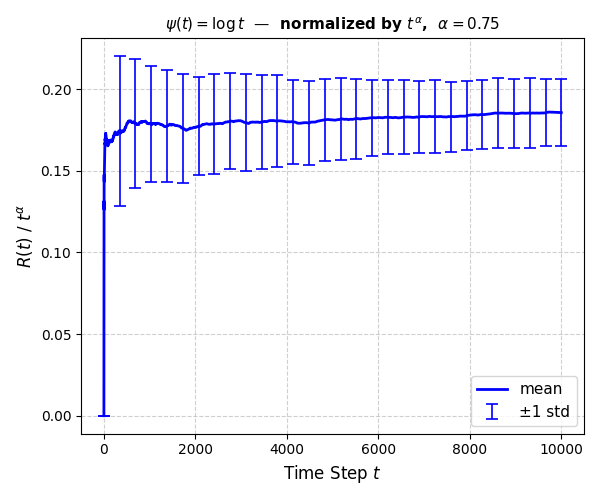}
    \caption[Normalized regret vs time plot showing $\Theta(t^{0.75})$ regret (instance (0.4,0.2), $\psi(t)=\log t$)]{The figure shows the $\frac{R(t)}{t^{0.75}}$ vs $t$ plot with the generator $\psi(t) = \log t$ on the instance (0.4,0.2). The  plot flattens with time, showing that $R(t) = \Theta(t^{0.75})$ as expected.}
    \label{fig:13}
\end{figure}

\begin{figure}[H]
    \centering
    \includegraphics[scale=0.6]{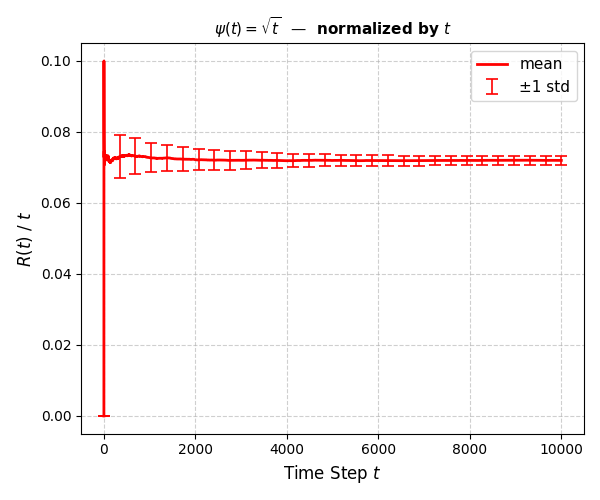}
    \caption[Normalized regret vs time plot showing $\Theta(t)$ regret (instance (0.6,0.4), $\psi(t)=\sqrt t$)]{The figure shows the $\frac{R(t)}{t}$ vs $t$ plot with the generator $\psi(t) = \sqrt t$ on the instance (0.6,0.4). The  plot flattens with time, showing that $R(t) = \Theta(t)$ as expected.}
    \label{fig:14}
\end{figure}

\begin{figure}[H]
    \centering
    \includegraphics[scale=0.6]{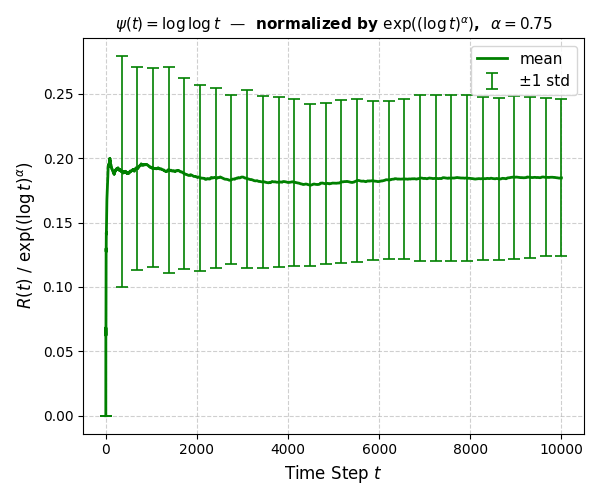}
    \caption[Normalized regret vs time plot showing $\Theta(\exp({(\log t)^{0.75}}))$ regret (instance (0.4,0.2), $\psi(t)=\log\log t$)]{The figure shows the $\frac{R(t)}{\exp({(\log t)^{0.75}})}$ vs $t$ plot with the generator $\psi(t) = \log \log t$ on the instance (0.4,0.2). The  plot flattens with time, showing that $R(t) = \Theta(\exp({(\log t)^{0.75}}))$ as expected.}
    \label{fig:15}
\end{figure}

\begin{figure}[H]
    \centering
    \includegraphics[scale=0.6]{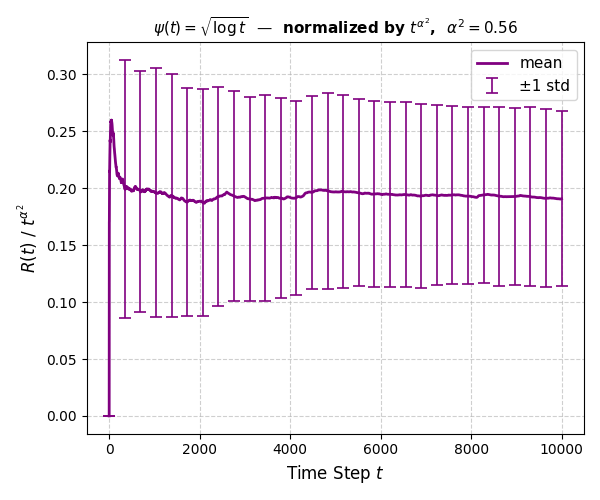}
    \caption[Normalized regret vs time plot showing $\Theta(t^{0.75^2})$ regret (instance (0.4,0.2), $\psi(t)=\sqrt{\log t}$)]{The figure shows the $\frac{R(t)}{t^{0.75^2}}$ vs $t$ plot with the generator $\psi(t) = \sqrt{\log t}$ on the instance (0.4,0.2). The  plot flattens with time, showing that $R(t) = \Theta(t^{0.75^2})$ as expected.}
    \label{fig:16}
\end{figure}

We plotted the normalized regret vs time for 4 instances: (0.4,0.6), (0.2,0.4), (0.5,0.75) and (0.8,0.2).  For each instance, we performed $100$ simulations for $10^4$ time steps. Figures \ref{fig:1}-\ref{fig:16} show the plots. It can be seen that the curves tend to plateau with time, indicating that the regret bound is achieved. We used a 16 core processor with 24 GB RAM, and the whole experiment took about 30 minutes to run.

\end{document}